\documentclass[journal]{article}
\usepackage{natbib}
\usepackage{amsmath}
\usepackage{amssymb}
\usepackage{multirow}
\usepackage{amsfonts}
\usepackage{mathrsfs}
\usepackage{booktabs}
\usepackage{authblk}
\usepackage{algorithm}
\usepackage{algorithmic}     
\usepackage{booktabs}
\usepackage{amsthm}
\usepackage{latexsym}
\usepackage{graphicx}
\usepackage{rotating}
\usepackage{subfig}
\usepackage{xcolor}
\usepackage{color,varioref}
\usepackage{longtable}
\allowdisplaybreaks[4]
\usepackage[colorlinks,linkcolor=blue]{hyperref}
\usepackage{diagbox} 

\newtheorem{theorem}{Theorem}
\newtheorem{assumption}{Assumption}
\newtheorem{definition}{Definition}
\newtheorem{lemma}{Lemma}
\newtheorem{remark}{Remark}

\usepackage{geometry}
\usepackage{makecell}
\usepackage{textcomp}

\newcommand{\comment}[1]{\hfill\textcolor{gray}{\texttt{//}\textit{#1}}}

\begin{document}


\title{Neural Network Training via Stochastic Alternating Minimization with Trainable Step Sizes}
	
\author{Chengcheng Yan\thanks{School of Mathematics and Computational Science, Xiangtan University, Xiangtan, 411105, China. Email: ycc956176796@gmail.com}, \hskip 0.2cm
Jiawei Xu\thanks{School of Mathematics and Computational Science, Xiangtan University, Xiangtan, 411105, China. Email: xujiawei@smail.xtu.edu.cn}, \hskip 0.2cm
Zheng Peng\thanks{School of Mathematics and Computational Science, Xiangtan University, Xiangtan, 411105, China. Email: pzheng@xtu.edu.cn},\hskip 0.2cm
Qingsong Wang\thanks{Corresponding author. School of Mathematics and Computational Science, Xiangtan University, Xiangtan, 411105, China. Email: nothing2wang@hotmail.com}}
	
\date{}	
\maketitle
\begin{abstract}
The training of deep neural networks is inherently a nonconvex optimization problem, yet standard approaches such as stochastic gradient descent (SGD) require simultaneous updates to all parameters, often leading to unstable convergence and high computational cost. To address these issues, we propose a novel method, \textbf{S}tochastic \textbf{A}lternating \textbf{M}inimization with \textbf{T}rainable Step Sizes (SAMT), which updates network parameters in an alternating manner by treating the weights of each layer as a block. By decomposing the overall optimization into sub-problems corresponding to different blocks, this block-wise alternating strategy reduces per-step computational overhead and enhances training stability in nonconvex settings. To fully leverage these benefits, inspired by meta-learning, we proposed a novel adaptive step size strategy to incorporate into the sub-problem solving steps of alternating updates. It supports different types of trainable step sizes, including but not limited to scalar, element-wise, row-wise, and column-wise, enabling adaptive step size selection tailored to each block via meta-learning. We further provide a theoretical convergence guarantee for the proposed algorithm, establishing its optimization soundness. Extensive experiments for multiple benchmarks demonstrate that SAMT achieves better generalization performance with fewer parameter updates compared to state-of-the-art methods, highlighting its effectiveness and potential in neural network optimization.

\end{abstract}
	
	
\maketitle
	
\section{Introduction}

Stochastic optimization forms the foundation of classical machine learning. Given a parameter set $\mathbf{W} \in \{W_l\}_{l=1}^{L}$ for the $L$-layer model, the typical goal is to solve the following stochastic optimization problem:
\begin{align}
\label{problem:1}
\min_{\mathbf{W}} F(\mathbf{W}) := \mathbb{E}_{ (\mathbf{x}, \mathbf{y}) \sim S } \left[ \mathcal{L} (\phi(\mathbf{x}; \mathbf{W}), \mathbf{y}) \right],
\end{align}
where $\mathcal{L}$ denotes the nonconvex loss function, $\mathbb{E}[\cdot]$ represents the expectation operator, and $\phi(\mathbf{x};\mathbf{W})$ denotes the model's predictive function parameterized by $\mathbf{W}$. Specifically, data pairs $(\mathbf{x}, \mathbf{y})$ represents data samples drawn from an unknown distribution $S = \{(x_i,y_i)\}_{i=1}^{N}$ defined by the train dataset, with $\mathbf{x} \in \mathbb{R}^d$ as the input and $\mathbf{y}$ as the ground truth.

However, it is well known that the above stochastic optimization problem is a complex non-convex problem. Taking the neural network model~\cite{lecun2015deep} as an example, e.g., multi-layer perceptron neural network~\cite{gardner1998artificial} (MLP) or convolutional neural network~\cite{gu2018recent} (CNN), the introduction of nonlinear layers coupled with high-dimensional parameters makes it generally intractable for the function $F(\mathbf{W})$ to find the global minimizer $\mathbf{W}^{*}$. Instead, a common practice is to seek a critical point $\mathbf{W}^{*}$ that satisfies $\nabla F(\mathbf{W}^{*}) = 0$. Fortunately, after introducing the stochastic sample gradient, the problem \eqref{problem:1} can be solved iteratively. This is the basic idea of stochastic gradient descent (SGD)~\cite{robbins1951stochastic}, which extends from the GD method~\cite{mason1999boosting}. It uses mini-batches to perform gradient descent on weight parameters and ultimately obtains updated weight values.

In this work, we define $f_i(\mathbf{W}) = \mathcal{L} (\phi(x_i; \mathbf{W}), y_i)$ for the $i$-th sample. We focus on the following nonconvex optimization problem:
\begin{align}
\label{problem:2}
\min_{\mathbf{W}} f(\mathbf{W}) := \frac{1}{N} \sum_{i=1}^{N} f_i(\mathbf{W}) .
\end{align}
Applying the mini-batch strategy, the corresponding iterative formula is given as follows:
\begin{align}
\label{eq:updatew}
    \mathbf{W}^{t+1} = \mathbf{W}^{t} - \eta^t \cdot \mathbf{g}_{t}  ,
\end{align}
where $\mathbf{g}_{t}$ is a stochastic gradient estimator computed over a mini-batch $B_t$, defined as $\mathbf{g}_{t} = \frac{1}{|B_t|} \sum_{i \in B_t} \nabla f_i(\mathbf{W}^t)$, here $\nabla f_i(\mathbf{W}^t)$ denotes the gradient of the $i$-th sample at the point $\mathbf{W}^t$, and $\eta^t \in (0,1)$ denotes the non-trainable step size, which is scalar. 

As mentioned previously, neural network training methods predominantly rely on the vanilla SGD algorithm and its variants, where all weight parameters are updated collectively through forward-backward propagation. However, for an $L$-layer neural network with weight $\mathbf{W} \in \{W_l\}_{l=1}^{L}$, \eqref{problem:2} can be equivalently reformulated as a multi-block alternating optimization problem by dividing the original problem into multiple optimization sub-problems. Under this condition, parameter updates can be performed in an alternating fashion by fixing the weights of all layers except the current layer and updating the weights of the current layer accordingly. In recent years, the alternating minimization (AM) method has garnered increasing attention in neural network training~\cite{taylor2016training, wang2019admm, wang2020toward, zeng2021admm, wang2022accelerated, wang2024badm, konevcny2016federated, ebrahimi2025aa, yan2025triple}. It formulates the parameter update of each layer as an independent optimization sub-problem, enabling alternating updates across layers. By decomposing the overall optimization problem into sub-problems corresponding to different blocks, these block-wise alternating strategies reduce per-step computational overhead and enhance training stability in nonconvex settings. Nevertheless, most existing AM-based methods for neural network optimization operate on full-batch data and gradients; they do not incorporate stochastic samples into the alternating optimization process. In this work, we integrate stochastic sampling into the AM framework, combining the strengths of alternating optimization with the efficiency benefits of stochastic sampling.

Beyond optimizing the updating framework for neural networks, an increasing number of studies~\cite{ruder2016overview} have focused on improving the optimization method of the weight parameter $w$ to enhance model accuracy. These improvements can be broadly categorized into two aspects: step-size optimization~\cite{kingma2014adam} and gradient-direction refinement~\cite{amari1998natural,dembo1982inexact}. In this article, we mainly focus on step size optimization. The fundamental principle behind step size optimization is to adjust the step size adaptively rather than relying on a fixed value. A brief overview of these methods will be provided in Section \ref{sec:related}. However, to the best of our knowledge, most existing step-size adaptive algorithms are built upon variants of the Adam algorithm, and their essential mechanism lies in scaling the initial step-size $\eta^{0}$ along different gradient directions; this process is not learned during training. Therefore, to fully leverage the benefits of AM methods, this study aims to update the weight parameters using trainable step sizes based on the AM strategy with increased training flexibility. In particular, the step size can be made trainable and can take various forms, including but not limited to scalar, element-wise, row-wise, and column-wise configurations. This flexibility allows the step-size type to be designed according to specific requirements.

\begin{figure}
\centering
\includegraphics[width=0.9\linewidth]{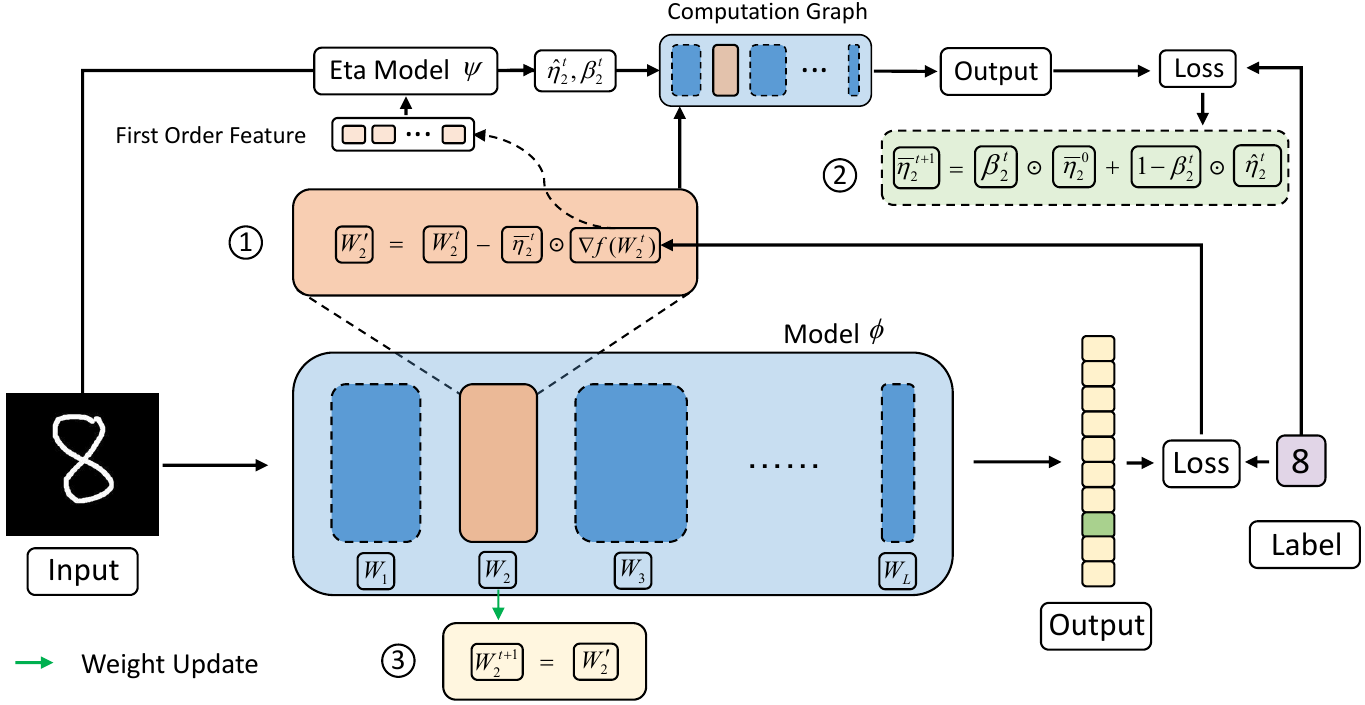}
    \caption{ The illustration describes the weight parameter update process of the SAMT method with Algorithm \ref{algo:learnable_nonscalar}. To obtain the updated weight $W_{2}^{t+1}$ of model $\phi$ in an alternating manner at \textcircled{3}, the algorithm first applies gradient descent to compute the temporary variable $W_{2}'$ at \textcircled{1}.  This intermediate result is then fed into the computational graph of the original model $\phi$ for forward inference, allowing the variable updating associated with the temporary value $\hat{\eta}_{2}^{t}$ and scale factor $\beta^{t}_t$ generated by the eta model $\psi$. Subsequently, the iterative formula at \textcircled{2} is employed to update trainable step size $\bar{\eta}_2^{t}$ (see Figure \ref{fig:stepsize}). For clarity, the stochastic sampling process is omitted from the illustration. } 
\label{fig:etann}
\end{figure}



In summary, the key contributions of this study can be summarized as follows:

\begin{itemize}
    \item We develop a novel method for training neural networks (see Figure \ref{fig:etann}), which comprises two key components: the stochastic AM framework and a strategy for the online trainable step sizes. By leveraging these two concepts, our algorithm enables adaptive updates of neural network weight parameters in an alternating process. In contrast to most existing stochastic algorithms, which update all weight parameters simultaneously, our approach capitalizes on the alternating optimization mechanism. This allows for the optimization of the current parameters utilizing the alternating update form, achieved with fewer parameter updates for better performance. Furthermore, we provide a general convergence proof in the alternating update process under mild conditions.

    \item We propose a novel strategy for trainable step sizes based on meta-learning. Specifically, our method takes five first-order features (i.e., the mean, variance, maximum, minimum, and norm values) of the gradient concerning the weight being updated as input to the eta model (see Figure \ref{fig:stepsize}), then outputs the final trainable step. This approach avoids the tedious process of manually calculating hyperparameter gradients~\cite{baydin2018hypergradient}. Notably, Distinct from most existing studies that primarily employ scalar step sizes, our method extends to a broader range of step size representations easily, including but not limited to scalar, element-wise, row-wise, and column-wise.

    \item Extensive experiments were conducted on multiple benchmark datasets. In particular, we designed experiments on MLP, CNN, and regression tasks, and conducted comparative evaluations using diverse datasets. Furthermore, to evaluate the generalization capability of the proposed algorithm, we performed hyperparameter sensitivity analyses and ablation studies. Overall, the proposed SAMT method outperforms existing algorithms in terms of test accuracy and robustness.

\end{itemize}

The rest of this paper is organized as follows: Section \ref{sec:related} provides a summary of recent related work. In Section \ref{sec:method}, we introduce the proposed SAMT algorithm. Section \ref{sec:converage} presents a detailed analysis of its convergence properties. Extensive experimental results on benchmark datasets are reported in Section \ref{sec:Experiments}, while Section \ref{sec:addition} presents additional comparative studies. Section \ref{sec:future} lists the limitations of the proposed algorithm and future work. Finally, Section \ref{sec:Conclusion} concludes the paper.

\section{Related Work}
\label{sec:related}

In recent years, with the widespread adoption of SGD~\cite{robbins1951stochastic} in deep learning, increasing research efforts have been dedicated to improving the efficiency of neural network parameter updates. These advancements primarily focus on refining the gradient direction and optimizing the step size within the gradient descent framework. Additionally, as an alternative to gradient-based methods, a class of algorithms based on AM methods has garnered increasing attention due to its theoretical convergence guarantees and innovative approach to solving sub-problems iteratively. In the following content, we will present a concise overview of these studies.

\begin{table}[thb]
\centering
\caption{Summary of the properties of the SAMT algorithm and several existing methods. ``-'' means not employed.}
\label{tab:summary}
\resizebox{1\textwidth}{!}{
\small
\begin{tabular}{c|c|c|c|c|c|c|c}
\toprule
\multirow{2.5}{*}{ \makecell[c]{Type} } & \multirow{2.5}{*}{ \makecell[c]{Algorithm} } & \multirow{2.5}{*}{ \makecell[c]{Stochastic} } & \multirow{2.5}{*}{ \makecell[c]{Trainable Step} } & \multirow{2.5}{*}{ \makecell[c]{Convergence Guarantee} }  & \multicolumn{3}{c}{Step Type} \\ 
\cmidrule{6-8}
 &  &  &  &  & Scalar & Element & Other \\
\midrule
\multirow{3}{*}{\makecell[c]{Non-AM}} & SGD \cite{robbins1951stochastic} & $\checkmark$ & - & $\checkmark$ & - & - \\
 & Adam \cite{kingma2014adam} & $\checkmark$ & - & $\checkmark$ & $\checkmark$ & $\checkmark$ & -\\
 & HD \cite{baydin2018hypergradient} & $\checkmark$ & $\checkmark$ & - & $\checkmark$ & $\checkmark$ & -\\
\midrule
\multirow{7}{*}{\makecell[c]{AM}} & scaleADMM \cite{taylor2016training} & - & - & $\checkmark$ & $\checkmark$ & - & - \\
 & BCD~\cite{zeng2019global} & - & - & $\checkmark$ & $\checkmark$ & - & -\\
 & dlADMM~\cite{wang2019admm} & - & - & $\checkmark$ & $\checkmark$ & - & - \\
 & mDLAM~\cite{wang2022accelerated} & - & - & $\checkmark$ & $\checkmark$ & - & - \\
 & Dante~\cite{sinha2020dante} & $\checkmark$ & - & - & $\checkmark$ & - & - \\
 & \textbf{SAMT (Our)} & $\checkmark$ & $\checkmark$ & $\checkmark$ & $\checkmark$ & $\checkmark$ & $\checkmark$ \\
\bottomrule
\end{tabular}
}
\end{table}

Table \ref{tab:summary} summarizes the key differences between the proposed algorithm and existing approaches. As shown, our method leverages alternating optimization strategies and updates model parameters using trainable step sizes. In addition to supporting both scalar and element-wise step sizes, the algorithm can easily extend to other variations of step size (e.g., column-wise and row-wise) as well. This highlights the flexibility of our approach, which is not constrained to a specific form of step size.

\subsection{Step Size Optimization}

For gradient-based optimization, an objective function is optimized by leveraging its gradients with respect to the model parameters. Beyond these standard gradients, a hypergradient concept is to be proposed for the derivative of the objective function with respect to the optimization procedure's hyperparameters, such as the learning rate, momentum, or regularization coefficients. Following this idea, Baydin et al. \cite{baydin2018hypergradient} introduced the concept of hypergradients concerning a scalar learning rate, enabling an adaptive update mechanism for step sizes based on hypergradients. This process is obtained through the chain differentiation rule. Subsequently, Chandra et al. \cite{chandra2022gradient} extended the work of \cite{baydin2018hypergradient} by introducing the utilize of automatic differentiation to compute hypergradients automatically. This approach allows the optimizer to dynamically update the associated hyperparameters, eliminating the need for manually deriving partial derivatives, thereby streamlining the optimization process. Retsinas et al. \cite{retsinas2023newton} formulate first and second-order gradients for automatically adjusting the learning rate during the training process, which is also the extended version of the hypergradient above. 

To the best of our knowledge, although the above excellent studies have successfully enabled step size adaptation via hypergradients, they inherently depend on manually designed differentiation chain rules. To address this limitation while incorporating the characteristics of alternating optimization, we propose a novel eta model $\psi$ to control the update of the step size. Specifically, as illustrated in Figure~\ref{fig:stepsize}, $\psi$ directly outputs the learnable parameters $\beta^{t}$ and $\hat{\eta}^{t}$, which govern the update dynamics. This design enhances the flexibility of the alternating update mechanism and eliminates the need for manual chain rule derivations. Furthermore, our approach allows the step size to be learned beyond scalar values. By leveraging dimensional expansion, our framework can be easily extended to support various forms of step sizes, such as scalar, element-wise, column-wise, row-wise, etc, configurations.

Certainly, in addition to utilizing hypergradient, there are also well-established adaptive learning rate methods that dynamically adjust the learning rate based on gradient changes. The primary algorithms in this category include AdaGrad~\cite{duchi2011adaptive}, RMSProp~\cite{tieleman2017divide}, Adam~\cite{kingma2014adam}, AMSGrad~\cite{reddi2018convergence}, Adabound~\cite{luoadaptive}, and other related methods. They accelerate algorithm convergence by incorporating momentum-based methods such as the Nesterov acceleration~\cite{nesterov1983method} and other momentum methods~\cite {kim2016optimized,polyak1964some}. While these methods can adjust the step size based on gradient information, they essentially scale a fixed initial step size $\eta^{0}$ along different gradient directions and do not support step size learnability. In contrast, our method updates the step size iteratively and enables online learnability. Furthermore, learning rate adjustment schedules, but not limited to decay and cosine annealing, among others, are commonly employed to refine the optimization process.

\subsection{AM-based Methods}
Although previous methods in terms of SGD are widely adopted in neural networks, gradient-based backpropagation often faces challenges in the vanishing gradient problem~\cite{hanin2018neural}. Fortunately, recent studies have proposed an alternative approach to address this issue by using an AM framework, which updates the network parameters in an alternating manner.  

Recent years have seen AM algorithms used extensively in deep learning, especially for handling non-convex optimization problems and tasks that demand iterative parameter updates \cite{taylor2016training,wang2019admm,gabay1976dual}. The optimization process of AM methods usually breaks down the original problem into multiple sub-problems, which are solved in an alternating manner during the whole update process. Following this concept, Taylor et al.~\cite{taylor2016training} proposed a novel training method employing alternating direction methods (scaleADMM) and Bregman iteration to train networks without gradient descent. Drawing inspiration from scaleADMM~\cite{taylor2016training}, Wang et al.~\cite{wang2019admm} proposed a novel optimization framework for deep learning based on the ADMM method~\cite{gabay1976dual} (dlADMM) to address the lack of slow convergence rates and global convergence guarantees. To enhance the efficiency of the algorithm's iterative updates, Wang et al.~\cite{wang2020toward} proposed a parallel deep learning ADMM framework (pdADMM) to achieve layer parallelism, which allows parameters of each neural network layer to be updated independently in a parallel manner. To evaluate the impact of activation functions and batch size on the performance of neural networks, Zeng et al.~\cite{zeng2021admm} developed an ADMM-based deep neural network training method that reduces the saturation problem of sigmoid-type activations. In addition, Wang et al. expanded the work of dlADMM~\cite{wang2019admm} by employing an inequality-constrained formulation to approximate the original problem, and utilizing the Nesterov acceleration to facilitate convergence for the proposed mDLAM~\cite{wang2022accelerated} method. 

Lately, Wang et al.~\cite{wang2024badm} took advantage of the ADMM framework to develop a novel data-driven algorithm. The key idea is to partition the training data into batches, which are further subdivided into sub-batches for parameter updates via multi-level aggregation. This algorithm can be regarded as a specialized variant of federated learning~\cite{konevcny2016federated}, specifically designed for neural network training. For the acceleration of deep learning models, Zeinab et al. \cite{ebrahimi2025aa} proposed an accelerated framework for training deep neural networks, which integrates Anderson acceleration within an AM approach inspired by ADMM to tackle the slow optimization problem. Yan et al. \cite{yan2025triple} proposed a triple-inertial accelerated alternating minimization (TIAM) framework for neural network training, which incorporates a triple-inertial acceleration strategy to speed the coverage of the neural model.

Drawing upon these outstanding prior works, this study incorporates stochastic algorithms into the AM framework. It explores the performance of neural network models under this setting by utilizing trainable step sizes. Concretely, in previous work, the application of AM algorithms in deep learning models has largely relied on full-batch training methods. To address this limitation, we integrate the advantages of stochastic gradients into the AM framework, thereby extending the applicability of AM methods based on stochastic gradients in neural network training. Furthermore, different from previous studies, which primarily employed a scalar step size for model training, we explored different types of learnable step sizes for the AM procedure.

\section{Trainable Step Size for Stochastic AM Method}
\label{sec:method}

In this section, we present a detailed introduction to our proposed SAMT algorithm, focusing on two key problems: leveraging stochastic samples for alternating updates of the neural network and learning trainable step sizes in an online manner within the AM framework. We then integrate these two components to develop a stochastic AM method with trainable step sizes. To ensure notational clarity, we adopt the following conventions: the symbol $\| \cdot \|$ is defined as the $\ell_2$ norm $\| \cdot \|_2$ for vector inputs, and as the Frobenius norm $\| \cdot \|_F$ for matrix inputs. Moreover, $\nabla f(x)$ denotes the gradient $\frac{\partial f(x)}{\partial x}$, and $\bar{\eta}$ denotes the trainable step size.

\subsection{Stochastic Alternating Optimization for Updating Network Parameters} 

As the name implies, the alternating optimization method refers to the process of optimizing parameters by iteratively fixing a subset of network parameters while updating a specific parameter at each step. Considering a neural network with $L$ layers, the architecture usually consists of two fundamental components: (i) the linear mapping layer $\mathbf{W}= \{W_l\}_{l=1}^L$ denotes the set of all weights throughout the network, and (ii) the activation function $\sigma:\mathbb{R}^{n}\to\mathbb{R}^{n}$. It is coupled and nested, forming the basic structure of a deep learning model. The output of the preceding mapping layer serves as the input to the activation layer of the subsequent layer, formulated as $\mathbf{a}_l=\sigma_l ( W_l \star \mathbf{a}_{l-1}) \in\mathbb{R}^{n}$, and $\mathbf{x} = \mathbf{a}_{0}$. In our work, inspired by the mathematical formulation of the optimization problem \eqref{problem:1}, a multi-layer neural network is constructed by nesting these layers, forming a composite prediction function $\phi$ defined as follows:
\begin{align}
    \phi(\mathbf{x} ; \mathbf{W}) =  \mathcal{F}_L \circ \sigma_{L-1} \circ \mathcal{F}_{L-1} \circ \cdots \circ \mathcal{F}_{1}(\mathbf{x}), \quad ( l = 1, \dots, L ),
\end{align}
where $\mathcal{F}_l = W_l \star (\cdot) $ denotes the general linear operator at $l$-th layer, and the operator $\circ$ denotes the composition of functions, applied from right to left. Specifically, $\star:=\cdot$ means the matrix multiplication for fully connected layers, and $\star:=*$ means the convolution operation for convolution layers. To enhance readability, we define $\mathcal{F}_l^{t}(\cdot) = W_l^{t} \star (\cdot)$ at the $t$-iteration. 

If we introduce stochastic samples $\{(x_i,y_i)\}_{i=1}^b$ from a distribution $S$ into the above formula, we can derive the following formulation based on problem \eqref{problem:2}:
\begin{align}
\label{eq:randomE}
    \min_{\mathbf{W}}\mathbb{E}_{(\mathbf{x},\mathbf{y})\sim\mathcal{S}}[\mathcal{L}(\phi( \mathbf{x}; \mathbf{W}),\mathbf{y})]  ,
\end{align}
where $\mathcal{L}$ denotes the loss function. However, a key question remains: how can the weight parameters of neural networks be alternately updated from the perspective of stochastic samples? Fortunately, similar to the update mechanism of block coordinate descent (BCD)~\cite{tseng2001convergence}, we treat each layer as a distinct block. When updating the parameters of a given block, the parameters of other blocks remain fixed. This strategy is consistently applied when updating different layers, ensuring that the algorithm systematically traverses and updates the parameters across all layers. 

Building on the aforementioned ideas, we now provide a detailed elaboration of the update process in the proposed method. Taking a three-layer neural network model as a case, it consists of three weight parameters (i.e., $W_1, W_2$, and $W_3$). To facilitate the analysis, we take a mean squared error (MSE) loss function as an example. Consequently, the optimization problem in \eqref{eq:randomE} can be reformulated as follows:
\begin{align}
    \min_{\mathbf{W}}\mathbb{E}_{(\mathbf{x},\mathbf{y})\sim\mathcal{S}}  \| \phi( \mathbf{x}; \mathbf{W}) - \mathbf{y} \|^{2}  ,
\end{align}
where $\phi( \mathbf{x}; \mathbf{W})$ = $\mathcal{F}_3 \circ \sigma_{2} \circ \mathcal{F}_{2} \circ \sigma_{1} \circ \mathcal{F}_{1}(\mathbf{x})$. To simplify the notation, we represent $\phi_{W_1}$ as a function in terms of $W_1$, with the other weight parameters held fixed. This implies that the remaining weight parameters are considered constant with respect to $W_1$. When optimizing a specific parameter, other unrelated parameters are treated as constants, which is a standard approach in optimization. Therefore, the process of alternating updates can be described as follows:
\begin{align}
    \label{process:am}
    W_1^{t} \leftarrow \arg\min_{W_1}\mathbb{E}_{(\mathbf{x},\mathbf{y})\sim\mathcal{S}}  \| \phi_{W_1}( \mathbf{x};  \cdot ) - \mathbf{y} \|^{2}  ,  \nonumber \\ 
    W_2^{t} \leftarrow \arg\min_{W_2}\mathbb{E}_{(\mathbf{x},\mathbf{y})\sim\mathcal{S}}  \| \phi_{W_2}( \mathbf{x};  \cdot ) - \mathbf{y} \|^{2}  ,   \\ 
    W_3^{t} \leftarrow \arg\min_{W_3}\mathbb{E}_{(\mathbf{x},\mathbf{y})\sim\mathcal{S}}  \| \phi_{W_3}( \mathbf{x};  \cdot ) - \mathbf{y} \|^{2}  ,    \nonumber
\end{align}
where $\phi_{W_1}( \mathbf{x}; \cdot ) = \mathcal{F}_3^{t} \circ \sigma_{2} \circ \mathcal{F}_{2}^{t} \circ \sigma_{1} \circ \mathcal{F}_{1}(\mathbf{x}) $, and $\mathcal{F}_3^{t}$ and $\mathcal{F}_2^{t}$ represent the corresponding weight $W_3^{t}$ and $W_2^{t}$ obtained from the $t$-th iteration of the optimization process, while the $W_1$ in $\mathcal{F}_{1}(\mathbf{x})$ means the weight to be updated. The definitions of the other functions, $\phi_{W_2}$ and $\phi_{W_3}$, following a similar formulation as $\phi_{W_1}$. Therefore, \eqref{process:am} can be seen as the simplified AM framework in our work, which enables the alternating optimization of weight parameters. Next, we will introduce a brief overview of the $\arg\min$ operation in our proposed method, which involves the concept of minimization in the AM framework. The detailed procedures for solving the $\arg\min$ sub-problems in our proposed method will be summarized in Section \ref{sec:onlinestepsize}, which includes Algorithm \ref{algo:learnable_scalar} for scalar trainable step sizes and Algorithm \ref{algo:learnable_nonscalar} for non-scalar trainable step sizes.

In the general stochastic gradient descent algorithm, it is well established that given a mini-batch of training data, the optimization objective is to minimize the expected error and update the weight parameter $\mathbf{W}$. Due to the complexity of obtaining analytical solutions for updating individual neural network weight parameters, attributed to factors such as non-convexity and coupling, gradient descent is commonly employed to approximate the true solution iteratively. In our work, as shown in Algorithm \ref{algo:our}, our approach formulates the update of each weight parameter as an individual sub-problem. These sub-problems are solved iteratively using stochastic gradient descent to obtain the updated parameter values.

When solving each sub-problem, our algorithm will select data by sampling $\{(\mathbf{x}_i,\mathbf{y}_i)\}_{i=1}^b\sim\mathrm{Uniform}(S)$, where $b$ denotes the mini-batch size. Accordingly, the gradient can be computed based on the loss function with respect to the corresponding parameters. For ease of analysis, we denote $f(\mathbf{w})$ as the loss function value corresponding to a mini-batch with respect to a certain weight parameter $\mathbf{w}$ (e.g., $W_1, W_2, \mathrm{~or~} W_3$). Consequently, the weight values are then updated using the conventional gradient descent algorithm, respectively. The definition of $f(\mathbf{w})$ is given as follows:
\begin{align*}
    f_t(\mathbf{w}) \leftarrow \frac{1}{b}\sum_{i=1}^b(\mathbf{y}_i- \phi ( \mathbf{x}_i ; \mathbf{w} ) )^2   ,
\end{align*}
by subsequently applying the gradient descent algorithm, we obtain
\begin{align}
\label{eq:gd}
    \mathbf{w}^{t+1} \leftarrow \mathbf{w}^{t} - \eta^{t} \cdot {\mathbf{g}}_t  ,
\end{align}
where $\mathbf{g}_{t}$ is a stochastic gradient estimator computed over a mini-batch $B_t$, here $\mathbf{g}_{t} = \frac{1}{|B_t|} \sum_{i \in B_t} \nabla f_i(\mathbf{w}^t)$, and the hyperparameter $\eta^{t}$ represents the scalar step size associated with $\mathbf{w}$. It is worth emphasizing that, for the sake of clarity in our exposition, we have treated each layer as an individual block and performed alternating updates accordingly. 

In the preceding discussion, we outlined the alternating optimization process of the proposed method. The approach is relatively straightforward but effective, while \eqref{eq:gd} can be further extended by refining the step size $\eta$. In the next section, we will explore how trainable step sizes can be leveraged further to enhance the stability of our alternating optimization method.

\begin{figure}
\centering
\includegraphics[width=0.9\linewidth]{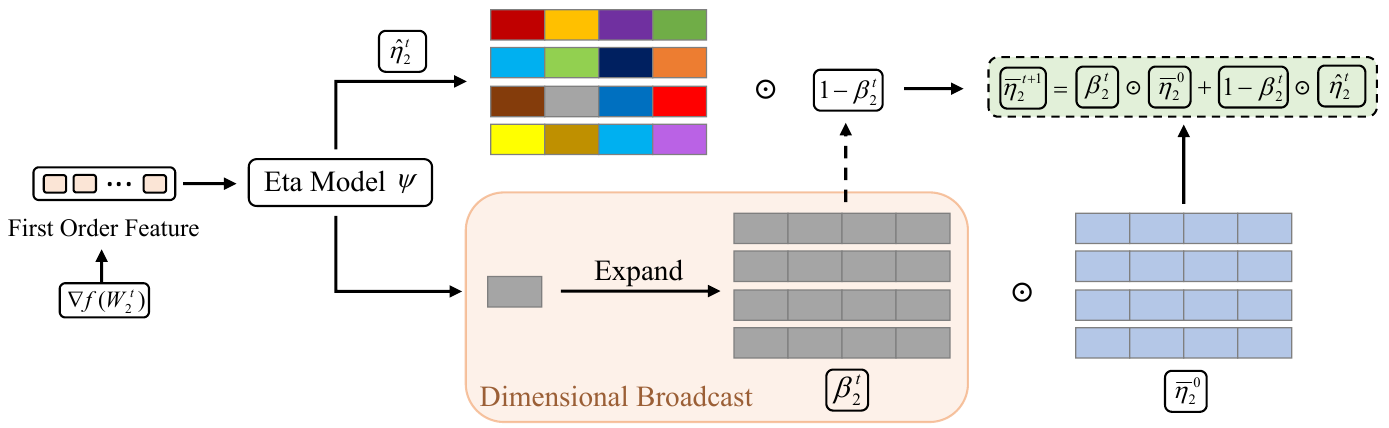}
    \caption{ The update process of Step \textcircled{2} in Figure~\ref{fig:etann}. Gradient-related five first-order feature vector (i.e, the mean, variance, maximum, minimum, and norm values) of $\nabla f(W_2^t)$ is first fed into the eta model $\psi$, which outputs a trainable step size $\hat{\eta}^{t}_2$ and a scaling factor $\beta^{t}_2$. These outputs are then combined with the initial step size $\bar{\eta}_2^{0}$ through a weighted summation to obtain the final step size $\bar{\eta}_2^{t+1}$, which is subsequently used to update the weight parameter $W_2$. In particular, different colors in the figure represent varying numerical values.} 
\label{fig:stepsize}
\end{figure}

\subsection{Online Update for Trainable Step Size} 
\label{sec:onlinestepsize}
In contrast to previous studies that predominantly focused on scalar step sizes, our work also includes investigating non-scalar step sizes. We contend that neural network weight parameters can be assigned distinct step sizes based on varying data input types, enabling the model to better adapt to diverse task requirements. For instance, in the classification task of birds, the head and wings in the upper part of the image are typically the key features used for distinguishing different species, while the leg information in the lower part is rarely utilized. This suggests that different tasks cannot rely solely on scalar step sizes to update the weight parameters. Instead, learnable step sizes should be assigned to different regions of the image, enabling online adaptive adjustments to the magnitude of parameter updates.

Specifically, to implement the learning of trainable step sizes, we leverage the concept of meta-learning~\cite{hospedales2021meta} and construct an eta model $\psi$ (i.e., a three-layer MLP model, default hidden number is 64) to compute the learnable values $\beta^t$ and $\hat{\eta}^{t}$, then to update the trainable step size $\bar{\eta}^t$, detailed update procedure can be seen in Figure \ref{fig:stepsize}. In contrast to existing methods that rely on hypergradient~\cite{baydin2018hypergradient}, our method depends on the output of the eta model $\psi$, which is integrated into the computation graph of the optimization model $\phi$, together with the updated temporary variable $W'$. Finally, the parameters $\beta^t$ and $\hat{\eta}^{t}$ are updated based on the resulting loss value. The whole procedure circumvents the computational complexity and manual derivation required by the chain rule. This enables a more efficient and flexible optimization process. Our approach can be readily extended to various types and shapes of trainable step sizes, rather than being limited to trainable scalar step sizes.

The update process and training of the eta model $\psi$ are a significant part. As illustrated in Figure \ref{fig:etann}, after computing the gradient information $\nabla f(W^{t})$ of the target weight parameter $W^{t}$, we do not directly feed the raw gradient into the eta model $\psi$. This is because, through experimental observation, we found that doing so leads to instability and oscillations in the learned step size during training. To mitigate this issue, we encode the original gradient $\nabla f(W^{t})$ using a set of first-order statistical feature vector, denoted as $\mathcal{D}^{t} = [d_1^{t}, d_2^{t}, \cdots, d_5^{t}]$, where each $d_i^{t}$ represents one of the following: the mean, variance, maximum, minimum, and norm of $\nabla f(W^{t})$. The eta model $\psi$ then takes $\mathcal{D}^{t}$ as input and produces the outputs $\beta^t$ and $\hat{\eta}^{t}$, so the training process of adaptive step size can be formulated as:
\begin{equation}
    \label{eq:min_etann}
    \min_{\beta,\hat{\eta}} \mathbb{E}_{ (\dot{\mathbf{x}}, \dot{\mathbf{y}}) \sim \dot{S} } \left[ \mathcal{L} (\phi(\dot{\mathbf{x}}; W'), \dot{\mathbf{y}}) \right],
\end{equation}
where temporary weight parameter $W' = W^{t} - \bar{\eta}^{t} \odot \nabla f(W^{t})$, scale factor and temporary step size $\beta^{t},\hat{\eta}^{t} \leftarrow \psi(\mathcal{D}^{t})$, here $\beta^{t} \in (0,1)$, and $(\hat{\eta}^{t})_{i} \in (0,1) , \forall i \in [1,size(\bar{\eta})]$. To enhance the robustness of the eta model $\psi$ and increase the diversity of training data, we sample the training instances for the eta model from the original distribution $S$ at intervals of two, resulting in a new subset denoted as $\dot{S}$, where $ (\dot{\mathbf{x}}, \dot{\mathbf{y}}) \sim \dot{S} \subseteq S$. Therefore, trainable step size $\bar{\eta}^{t}$ can be updated by following convex combination: 
\begin{equation}
\label{eq:update_etann}
    \bar{\eta}^{t+1} = \underbrace{ \beta^{t} \odot \bar{\eta}^{0} }_{\text{left term}} + \underbrace{ (1-\beta^{t}) \odot \hat{\eta}^{t} }_{\text{right term}}  ,
\end{equation}
where $\odot$ denotes the Hadamard Product, $\bar{\eta}^{0}$ denotes the initialization step size, and $\hat{\eta}^{t}$ is the output of the eta model $\psi$, which will be projected between 0 and 1 using our Tanh projection style. Comparative experiments and discussions of different projection styles will be analyzed in Section \ref{sec:projectionstyle}. The momentum-weighted convex combination update in \eqref{eq:update_etann} offers several advantages. It stabilizes training by retaining the initialization $\bar{\eta}^{0}$ while adaptively incorporating the current estimate $\hat{\eta}^{t}$. This strategy keeps $\bar{\eta}^{t+1}$ within a controlled range, preventing explosions and oscillations. The results of subsequent ablation studies further confirm this advantage.

As introduced previously, \eqref{eq:gd} represents a classical gradient descent algorithm with a scalar step size, where the weight parameters across all layers typically share a unified step size coefficient and are updated iteratively in this form. In this section, inspired by meta-learning, we extend $\eta$ to a trainable step size $\bar{\eta}$, including, but not limited to, different types of trainable step sizes: element-wise, row-wise, column-wise, and scalar (a special case of the element-wise where all elements share the same value). For example, in the case of element-wise, the procedures of our proposed trainable step size method are outlined as follows:

\begin{itemize}
    \item Initially, after executing backpropagation, we obtain the gradient of the stochastic sample, denoted as $\mathbf{g}_t = \frac{1}{|B_t|} \sum_{i \in B_t} \nabla f_i(\mathbf{w}^t)$. Utilizing the following update formula, ${\mathbf{w}}^{\prime} \leftarrow \mathbf{w}^{t} - \bar{\eta}^{t} \odot {\mathbf{g}}_t$, we compute the temporary weight parameter ${\mathbf{w}}^{\prime}$, where $\bar{\eta}^{t}$ represents the trainable step size at $t$-th iteration introduced previously.
    
    \item In the following stage, we constructed an eta model $\psi$ to generate a learnable scale factor $\beta^{t}$ and learnable step size $\hat{\eta}^{t}$ for updating $\bar{\eta}^{t}$,  The temporary weight parameters ${\mathbf{w}}^{\prime}$ were then forward-backward propagated through the computational graph (sharing the same architecture as the original model $\phi$) of the original model $\phi$. This process computes the gradients of the loss function with respect to $\beta^{t}$ and $\hat{\eta}^{t}$, enabling their subsequent updates. Combining the initialize step size $\bar{\eta}^{0}$, the update process of $\bar{\eta}^{t+1}$ follows from \eqref{eq:update_etann}.
    
    \item Concluding this process, using the gradient information derived in the previous step, we update the learnable parameters $\beta^{t}$ and $\hat{\eta}^{t}$ via standard gradient descent. The model parameters are then updated by setting $\mathbf{w}^{t+1} = \mathbf{w}^{\prime}$.
\end{itemize}

The whole procedure is detailed in Algorithm \ref{algo:learnable_nonscalar}, which corresponds to the non-scalar variant of our SAMT method. However, it should be noted that Algorithm~\ref{algo:learnable_scalar} can be viewed as a scalar version of Algorithm~\ref{algo:learnable_nonscalar}, where $\bar{\eta}$ reduces to a trainable scalar step size. For analytical convenience, in our convergence analysis, we primarily focuses on the convergence properties of the scalar version of Algorithm~\ref{algo:learnable_scalar}, as detailed in Section~\ref{sec:converage}.

\subsection{Stochastic Alternating Optimization Method with Scalar Trainable Step Size } 

Following the previous section, we combine the theories presented in the above two subsections, and using a three-layer neural network as an example, Algorithm \ref{algo:our} and Algorithm \ref{algo:learnable_scalar} provide a detailed description of the computational process of our proposed method. We begin by presenting the scalar version of our SAMT algorithm.

\begin{algorithm}[H]
\caption{Stochastic Alternating Optimization with Trainable Step Size (SAMT)}
\begin{algorithmic}[1]
\label{algo:our}
\STATE $\mathbf{Input:~}$ Max number of iterations $T_{MAX}$, initial values $W_3^0,W_2^0,W_1^0$. 
\STATE $t:=1$
\WHILE{ $t< T_{MAX}$ }
\STATE $W_1^{t} \leftarrow \arg\min_{W_1}\mathbb{E}_{(\mathbf{x},\mathbf{y})\sim\mathcal{S}}  \|   \mathcal{F}_3^{t-1} \circ \sigma_{2} \circ \mathcal{F}_{2}^{t-1} \circ \sigma_{1} \circ \mathcal{F}_{1}(\mathbf{x}) - \mathbf{y} \|^{2} $        \comment{With OAGD}
\STATE $W_2^{t} \leftarrow \arg\min_{W_2}\mathbb{E}_{(\mathbf{x},\mathbf{y})\sim\mathcal{S}}  \|   \mathcal{F}_3^{t-1} \circ \sigma_{2} \circ \mathcal{F}_{2} \circ \sigma_{1} \circ \mathcal{F}_{1}^{t} (\mathbf{x}) - \mathbf{y} \|^{2} $       \comment{With OAGD}
\STATE $W_3^{t} \leftarrow \arg\min_{W_3}\mathbb{E}_{(\mathbf{x},\mathbf{y})\sim\mathcal{S}}  \|   \mathcal{F}_3 \circ \sigma_{2} \circ \mathcal{F}_{2}^{t} \circ \sigma_{1} \circ \mathcal{F}_{1}^{t}(\mathbf{x}) - \mathbf{y} \|^{2} $       \comment{With OAGD}
\STATE $t:=t+1$
\ENDWHILE
\STATE $\mathbf{Output:~}$ $W_1^{t},W_2^{t},W_3^{t}$
\end{algorithmic}
\end{algorithm}

Algorithm \ref{algo:our} constitutes the core procedure of our approach, which decomposes the weight update process of the neural network into multiple sub-problems solved by the OAGD algorithm. It fixed other weight parameters to optimize the current weight parameters in an alternating manner, iteratively. In this work, we consider two versions of the OAGD method, corresponding to scalar and non-scalar trainable step sizes. Specifically, the SAMT-S variant employs the procedure outlined in Algorithm \ref{algo:learnable_scalar}, whereas other SAMT variants with different forms of step sizes adopt Algorithm \ref{algo:learnable_nonscalar} to solve each subproblem.

For the detailed resolution of each sub-problem, Algorithm \ref{algo:learnable_scalar} provides a description of the implementation of trainable scalar step sizes and the application of stochastic gradient descent for updating the current weight parameters, which corresponds to our proposed SAMT-S method. Furthermore, Section \ref{sec:converage} provides the convergence guarantee for the SAMT algorithm under the trainable scalar setting.

\begin{algorithm}[H]
\caption{Online Adaptive Gradient Descent with \textbf{Scalar} (OAGD-S)}
\begin{algorithmic}[1]
\label{algo:learnable_scalar}
\STATE $\mathbf{Input:~}$ Training data $S = \{(\mathbf{x}_i,\mathbf{y}_i)\}_{i=1}^N \in \mathbb{R}^d\times\mathbb{R}$, mini-batch size $b$, initial parameters $\mathbf{w}$ of model $\phi$, eta model $\psi$, trainable step size $\bar{\eta}$, and initial step size $\bar{\eta}^{0}$.
\FOR{$t=1$ to $T$}
\STATE Select data by sampling $\{(\mathbf{x}_i,\mathbf{y}_i)\}_{i=1}^b\sim\mathrm{Uniform}(S)$
\STATE $f_t(\mathbf{w}) \leftarrow \frac{1}{b}\sum_{i=1}^b(\mathbf{y}_i- \phi ( \mathbf{x}_i ; \mathbf{w} ) )^2$
\STATE $\mathbf{g}_t \leftarrow \nabla f_t(\mathbf{w}^t)$  
\STATE ${\mathbf{w}}^{\prime} \leftarrow \mathbf{w}^{t} - \bar{\eta}^{t} \cdot {\mathbf{g}}_t $         \comment{Parameter Update for Original Model $\phi$}
\STATE $ \beta^{t},\hat{\eta}^{t} \leftarrow \psi(\mathcal{D}^{t})$     
    \comment{The output of eta Model $\psi$}
\STATE $ \bar{\eta}^{t+1} = \beta^{t} \cdot \bar{\eta}^{0} + (1-\beta^{t}) \cdot \hat{\eta}^{t} $          \comment{Update Trainable Step Size $\bar{\eta}$}
\STATE $\mathbf{w}^{t+1} = \mathbf{w}^{\prime}$ \comment{Update Weight}
\ENDFOR
\STATE $\mathbf{Output:~}$ $\mathbf{w}^{T}$
\end{algorithmic}
\end{algorithm}

\subsection{Deep Alternating Optimization Method with Non-Scalar Trainable Step Size}

Although the combination of trainable scalar step sizes and alternating optimized weights can be applied to most of the model training, as previously discussed, assigning different types of step sizes to different parts of the input data is more aligned with the principle of feature selection. Step sizes tailored to specific input components allow for more diverse and fine-grained adaptations, thereby enhancing the optimization of model parameters. To this end, we extend the scalar step size formulation in Algorithm \ref{algo:learnable_scalar}  and propose a novel online adaptive gradient descent method that employs non-scalar, trainable step sizes, as shown in Algorithm \ref{algo:learnable_nonscalar}. Different from previous algorithms based on hypergradient or adaptive step size, this proposed algorithm is capable of autonomously learning different types of step sizes according to their designated types. It leverages a broadcast mechanism to adapt the step sizes flexibly, rather than being restricted to a predetermined shape.

Algorithm \ref{algo:learnable_nonscalar} outlines the implementation of our non-scalar adaptive gradient descent method. To formally characterize the role of non-scalar learnable step sizes in the optimization process, we consider the following update rule:
$$\mathbf{w}^t = \mathbf{w}^{t-1} - \bar{\eta}^t \odot \nabla f(\mathbf{w}^{t-1}) ,$$
where $\bar{\eta}^t$ denotes a trainable, non-scalar step size that is applied through the Hadamard (element-wise) product $\odot$. Notably, $\bar{\eta}^t$ is implemented from learnable parameters by using PyTorch's broadcasting mechanism, allowing for flexible step size type. It represents that $\bar{\eta}^t$ can take on different types depending on the application scenario and will be automatically broadcast to match the dimensionality of the gradient $\nabla f(\mathbf{w}^{t-1})$. This implementation enables a flexible strategy for adapting step sizes at various levels of granularity; the procedure for updating the trainable step size $\bar{\eta}^{t}$ is depicted in Figure~\ref{fig:stepsize}, in which we mainly focus on the proposed SAMT-E method under element-wise type step sizes.

\begin{algorithm}[H]
\caption{Online Adaptive Gradient Descent with \textbf{Non-Scalar} (OAGD-NS)}
\begin{algorithmic}[1]
\label{algo:learnable_nonscalar}
\STATE $\mathbf{Input:~}$ Training data $S = \{(\mathbf{x}_i,\mathbf{y}_i)\}_{i=1}^N \in \mathbb{R}^d\times\mathbb{R}$, mini-batch size $b$, initial parameters $\mathbf{w}$ of model $\phi$, eta model $\psi$, trainable step size $\bar{\eta}$, and initial step size $\bar{\eta}^{0}$.
\FOR{$t=1$ to $T$}
\STATE Select data by sampling $\{(\mathbf{x}_i,y_i)\}_{i=1}^b\sim\mathrm{Uniform}(S)$
\STATE $f_t(\mathbf{w}) \leftarrow \frac{1}{b}\sum_{i=1}^b(\mathbf{y}_i- \phi ( \mathbf{x}_i ; \mathbf{w} ) )^2$
\STATE $\mathbf{g}_t \leftarrow \nabla f_t(\mathbf{w}^t)$  
\STATE ${\mathbf{w}}^{\prime} \leftarrow \mathbf{w}^{t} - \bar{\eta}^{t} \odot {\mathbf{g}}_t $         \comment{Parameter Update for Original Model $\phi$}
\STATE $ \beta^{t},\hat{\eta}^{t} \leftarrow \psi(\mathcal{D}^{t})$     
    \comment{The output of eta Model $\psi$}
\STATE $ \bar{\eta}^{t+1} = \beta^{t} \odot \bar{\eta}^{0} + (1-\beta^{t}) \odot \hat{\eta}^{t} $          \comment{Update Trainable Non-Scalar Step Size $\bar{\eta}$}
\STATE $\mathbf{w}^{t+1} = \mathbf{w}^{\prime}$ \comment{Update Weight}
\ENDFOR
\STATE $\mathbf{Output:~}$ $\mathbf{w}^{T}$
\end{algorithmic}
\end{algorithm}

\section{Convergence Analysis}
\label{sec:converage}

For the sake of analytical traceability, in this section, we will explore the convergence guarantee for the optimization process of the proposed method under a scalar trainable step size. This chapter is primarily divided into two parts: preliminary concepts and convergence properties.

\subsection{Fundamentals}

Let $\{W_1^{*},W_2^{*}, \dots,W_L^{*}\}$ denote the global optimum of $f$ computed on the entire data population. For theoretical analysis, we assume that the algorithm knows the lower-bound on the radii of convergence $r_1,r_2,\dots,r_L$ for $W_1, W_2,\dots, W_L$.\footnote{This assumption is potentially easy to eliminate with a more careful choice of the step size in the first iterations.} Let $\nabla_if^1$ denote the gradient of $f$ computed for a single data sample $(x, y)$ and taken with respect to the $i$-th argument of the function $f$ (weights from Algorithm~\ref{algo:our}). 

In the next stage, we define $\nabla_i f(\cdot)$ as the gradient of $f$ concerning ${W_i}$ computed for the entire data population, i.e., an infinite number of samples (``oracle gradient''). We assume in the $i$-th step ($i = 1,2,\dots,L$), the AM algorithm follows the update rule:

\begin{equation}
\label{eq:update}
W_i^{t+1} \!=\! \Pi_i(W_i^t - \bar{\eta}^{t} \nabla_if^1(W_1^{t+1},\dots,W_{i-1}^{t+1},W_{i}^{t}, W_{i+1}^{t},\dots,W_L^t)),
\end{equation}
where $t$ denotes iteration, $\Pi_i$ denotes the projection onto the Euclidean ball $B_2(\frac{r_i}{2},W_i^0)$ of some given radius $\frac{r_i}{2}$ centered at the initial iterate weight parameter $W_i^0$. Thus, given any initial vector $W_i^{0}$ in the ball of radius $\frac{r_i}{2}$ centered at $W_i^{*}$, we are guaranteed that all iterates remain within an $r_i$-ball of $W_i^{*}$. This is true for all $i = 1,2,\dots, L$. The re-projection step of \eqref{eq:update} represents that
starting close enough to the optimum and taking small steps leads to the convergence rate of Theorem~\ref{theorem:final}. The radii dictate how convergence is affected if the iterates stray further from the optimum through the variable $\sigma^2$ (see \eqref{eq:mid000}) defined before that theorem.

\begin{remark}
Several differences can be identified between the AM scheme studied and our proposed Algorithm~\ref{algo:our}, as summarized below:
\begin{itemize}
    \item The AM scheme analyzed performs only a single stochastic gradient descent (SGD) step for updating the weight parameters at each iteration, whereas Algorithm~\ref{algo:our} allows full optimization of all weight parameters until convergence.
    \item The gradient direction is estimated based on a single data sample, while Algorithm~\ref{algo:our} employs mini-batch gradients in practice for improved stability and generalization.
    \item A re-projection step is incorporated in the AM scheme, which is not included in Algorithm~\ref{algo:our}.
\end{itemize}


It can be inferred that the generalized AM scheme analyzed yields \textbf{worst-case theoretical guarantees} relative to the original formulation in Algorithm~\ref{algo:our}. Specifically, we expect the convergence rate for the original setting in Algorithm~\ref{algo:our} to be no worse than the one dictated by the obtained statistical guarantees. This is attributed to the fact that, under the generalized AM framework, each variable undergoes a single stochastic update based on a single data point-while other variables remain fixed. In contrast, Algorithm~\ref{algo:our}, along with similar approaches in the literature, allows for increased mini-batch sizes at each AM step (i.e., a semi-stochastic regime), which typically leads to improved convergence properties. Moreover, by restricting each AM subproblem to be executed multiple times, we effectively analyze a more stochastic and potentially noisier variant of parameter updates.

\end{remark}


\subsection{ Assumptions and Analysis }
To facilitate the introduction of the convergence properties of the proposed algorithm, this subsection first presents key definitions and assumptions from optimization theory, which will serve as the theoretical foundation for the subsequent proofs. At the beginning, let $\mathcal{J} ( W_1, W_2, \dots,W_L) = -f( W_1, W_2, \dots,W_L)$ and define $\mathcal{J}^*_d(W_d)$ as the objective function with all optimal variables except $W_d$, i.e.,
$$\mathcal{J}^*_d(W_d)=\mathcal{J}( W_1^{*},W_2^{*},\dots,W_{d-1}^{*},W_d,W_{d+1}^{*},\dots,W_{L-1}^{*},W_L^{*}).$$

Let $\Omega_1,\Omega_2,\dots,\Omega_L$ denote non-empty compact convex sets such that for any $i = \{1,2,\dots,L\}, W_i\in\Omega_i$. The following assumptions are provided on $\mathcal{J}^*_d(W_d)$ ($d = 1,2,\dots,L$) and the objective function $\mathcal{J} ( W_1, W_2, \dots,W_L)$. 


\begin{assumption}[{\em Strong concavity}]
\label{assu:sc}
The function $\mathcal{J}^*_d(W_d)$ is strongly concave for all pairs $( W_{d,1}, W_{d,2})$ in the neighborhood of $ W_d^{*}$. That is
\begin{eqnarray*}
\mathcal{J}^*_d(W_{d,1}) - \mathcal{J}^*_d(W_{d,2}) - \left<\nabla_d \mathcal{J}^*_d(W_{d,2}), W_{d,1}-W_{d,2}\right>  \leq -\frac{\lambda_d}{2}\|W_{d,1} - W_{d,2}\|^2,
\end{eqnarray*}
where $\lambda_d > 0$ is a scalar value. 
\label{def:strongconN}
\end{assumption}

\begin{assumption}[{\em Smoothness}]
\label{assu:smooth}
The function $\mathcal{J}^*_d(W_d)$ is $\mu_d$-smooth for all pairs $(W_{d,1},W_{d,2})$. That is
\begin{eqnarray*}
\mathcal{J}^*_d(W_{d,1}) - \mathcal{J}^*_d(W_{d,2}) - \left<\nabla_d \mathcal{J}^*_d(W_{d,2}), W_{d,1}-W_{d,2}\right> \geq  -\frac{\mu_d}{2}\|W_{d,1} - W_{d,2}\|^2,
\end{eqnarray*}
where $\mu_d > 0$ is the smoothness constant. 
\label{def:smoothN}
\end{assumption}


\begin{assumption}[{\em Gradient stability (GS)}]
\label{assu:gs}
We assume $\mathcal{J}(W_1,W_2\dots,W_L)$ satisfies GS ($\gamma_d$) condition, where $\gamma_d \geq 0$, over Euclidean balls $W_1 \in B_2(r_1,W_1^{*}), \dots, W_{d+1}\in B_2(r_{d+1},W_{d+1}^{*}),\dots, W_{L} \in B_2(r_{L},W_{L}^{*})$ of the form
\vspace{-0.07in}
\[ 
\|\nabla_d \mathcal{J}^*_d(W_d) -\nabla_d \mathcal{J}(W_1, W_2,\dots,W_L)\|\leq \gamma_d\sum_{\substack{i=1 \\ i\neq d}}^L\|W_i - W_i^{*}\|.
\]
\vspace{-0.2in}
\label{def:GSN}
\end{assumption}

\begin{assumption}
    \label{assu:bounder}
    According to the iteration process of $\bar{\eta}^{t}$ follows that $\bar{\eta}^{t+1} = \beta^{t} \cdot \bar{\eta}^{0} + (1-\beta^{t}) \cdot \hat{\eta}^{t} $, here $ \beta^{t},\hat{\eta}^{t} \leftarrow \psi(\mathcal{D}^{t})$, for any $t$, there exits lower bounder $B_l$ and upper bounder $B_u = \frac{1}{ \gamma(L-1) } $ satisfies that $0 < B_l < B_u$ follows that 

    $$ 0 < B_l \leq  \bar{\eta}^{t}_{min}  \leq \bar{\eta}^{t} \leq  \bar{\eta}^{t}_{max} \leq B_u ,$$
where both $\beta^{t}$ and $\hat{\eta}^{t}$ are outputs of the eta model $\psi(\mathcal{D}^{t})$, satisfying $\beta^{t} \in (0,1)$ and $\hat{\eta}^{t} \in (0,1)$.

\end{assumption}


Based on the preceding theory, we present the following convergence guarantee theorem. The theoretical analysis presented in this work generalizes recent results on the statistical guarantees of the EM algorithm~\cite{balakrishnan2017statistical} to the broader AM framework. Under Assumptions \ref{assu:sc}, \ref{assu:smooth}, \ref{assu:gs}, and \ref{assu:bounder}, the following theorem provides a recursive formula for the expected error at each iteration of Algorithm~\ref{algo:our} under Algorithm \ref{algo:learnable_scalar} with scalar trainable size:

\begin{theorem}
\label{theorem:final}
Given the stochastic AM gradient iterates of the version of Algorithm~\ref{algo:our} under Algorithm \ref{algo:learnable_scalar} with scalar trainable size $\bar{\eta}^{t}$, the error at iteration $t+1$ satisfies
\begin{equation}
\begin{aligned}
\mathbb{E}\left[\sum_{d=1}^L\|\Delta^{t+1}_d\|^2\right] &\leq O(e^{-ct}) + O(1),
\end{aligned}
\end{equation}
where $\Delta^{t+1}_d := W_d^{t+1} - W_d^{*}$ for $d = 1,2,\dots,L$, and $c$ represent a constant value. For completeness, the proof of Theorem \ref{theorem:final} is presented in Appendix \hyperref[proof:final]{A}.

\end{theorem}


\section{Experimental Results}
\label{sec:Experiments}
Similar to the experimental setup of existing algorithms, to ensure a fair comparison, we constructed neural networks with different layer configurations to evaluate the test performance of various algorithms on neural networks.

\subsection{Experimental Settings and Methods }

In the specific experimental setup, we employed an MLP~\cite{gardner1998artificial} for grayscale single-channel images and a CNN~\cite{gu2018recent} for color three-channel images, respectively\footnote{The implementation will be publicly accessible at \url{https://github.com/yancc103/SAMT } once this manuscript is accepted.}. The selected loss function is cross-entropy with softmax, and all experimental train batches and test batches are set to 64 and 1000, respectively. By default, all algorithms employ the LeakyReLU~\cite{maas2013rectifier} activation function. The experiments were implemented in Python 3.8 with Torch 1.9~\cite{paszke2017automatic} and conducted on a Ubuntu 20.04.6 computer equipped with an Intel(R) Core(TM) i7-12700 2.10 GHz CPU and 16GB of memory.

For comparative experiments, we assess the proposed algorithm in relation to state-of-the-art optimization methods. Concretely, we compare it with stochastic gradient descent approaches, such as SGD~\cite{robbins1951stochastic}, Adam~\cite{kingma2014adam}, and the HD algorithm~\cite {baydin2018hypergradient} based on the hypergradient concept, as well as AM-based algorithms, including dlADMM~\cite{wang2019admm}, mDLAM~\cite{wang2022accelerated}, and Dante~\cite{sinha2020dante}. As specified in the original paper, the Dante algorithm adopts an alternating update strategy, performing five iterations of weight updates on a given layer before proceeding to the next. The stochastic gradient algorithm is implemented using PyTorch's standard optimizer, while the other algorithms are implemented with reference to the relevant code available in the literature. In our comparative experiments, we employ the best initial learning rate to evaluate the stochastic gradient algorithm. For AM-based algorithms, we adopt the optimal settings reported in their respective papers to ensure a fair comparison.

\subsection{Datasets}
We primarily utilize the following five datasets for the image classification task, which can be categorized into two main types: grayscale images and color images. The key differences among them are as follows: \\

\noindent\textbf{(1) Grayscale Images}
\begin{itemize}
    \item MNIST \cite{lecun1998gradient}: This dataset consists of 70,000 grayscale handwritten digit images of size $28\times28$, spanning 10 categories (0–9). It serves as a benchmark for handwritten digit recognition.
    \item Kuzushiji-MNIST (KMNIST) \cite{clanuwat2018deep}: Similar in dataset size and image resolution to the previous two, KMNIST features 10 categories of handwritten Japanese hiragana characters. It is designed for research on handwritten character recognition, particularly for non-Latin scripts.
    \item Fashion-MNIST (FMNIST) \cite{xiao2017fashion}: Sharing the same size and structure as MNIST, this dataset contains images of clothing items across 10 categories (e.g., T-shirts, shoes, bags). As a more challenging alternative to MNIST, it is particularly suitable for fashion classification tasks.
\end{itemize}

\noindent\textbf{(2) Color Images}
\begin{itemize}
    \item CIFAR-10 \cite{krizhevsky2009learning}: This dataset comprises 60,000 color images of size 32×32, distributed across 10 categories (e.g., airplanes, cars, cats, and dogs). It is widely used for general object classification tasks and presents a greater challenge compared to MNIST.
    \item CIFAR-100 \cite{krizhevsky2009learning}: With the same dataset size and image resolution as CIFAR-10, CIFAR-100 extends the number of categories to 100, making it suitable for fine-grained image classification, small-sample learning, and more complex model evaluations.
\end{itemize}

We maintained the same dataset configuration for all comparative experiments, and all of these datasets are readily available in PyTorch's official torchvision library~\cite{paszke2017automatic}.

\subsection{Comparison Experiments}
With reference to the experimental setup described above, this section provides a detailed analysis of our experimental results. As previously mentioned, for the image classification task, the five datasets were categorized into two groups of experiments for the MLP and CNN neural network structures for the grayscale and color image datasets, respectively.

\subsubsection{MLP Experiments}

\begin{figure}
\centering
\includegraphics[width=1.0\linewidth]{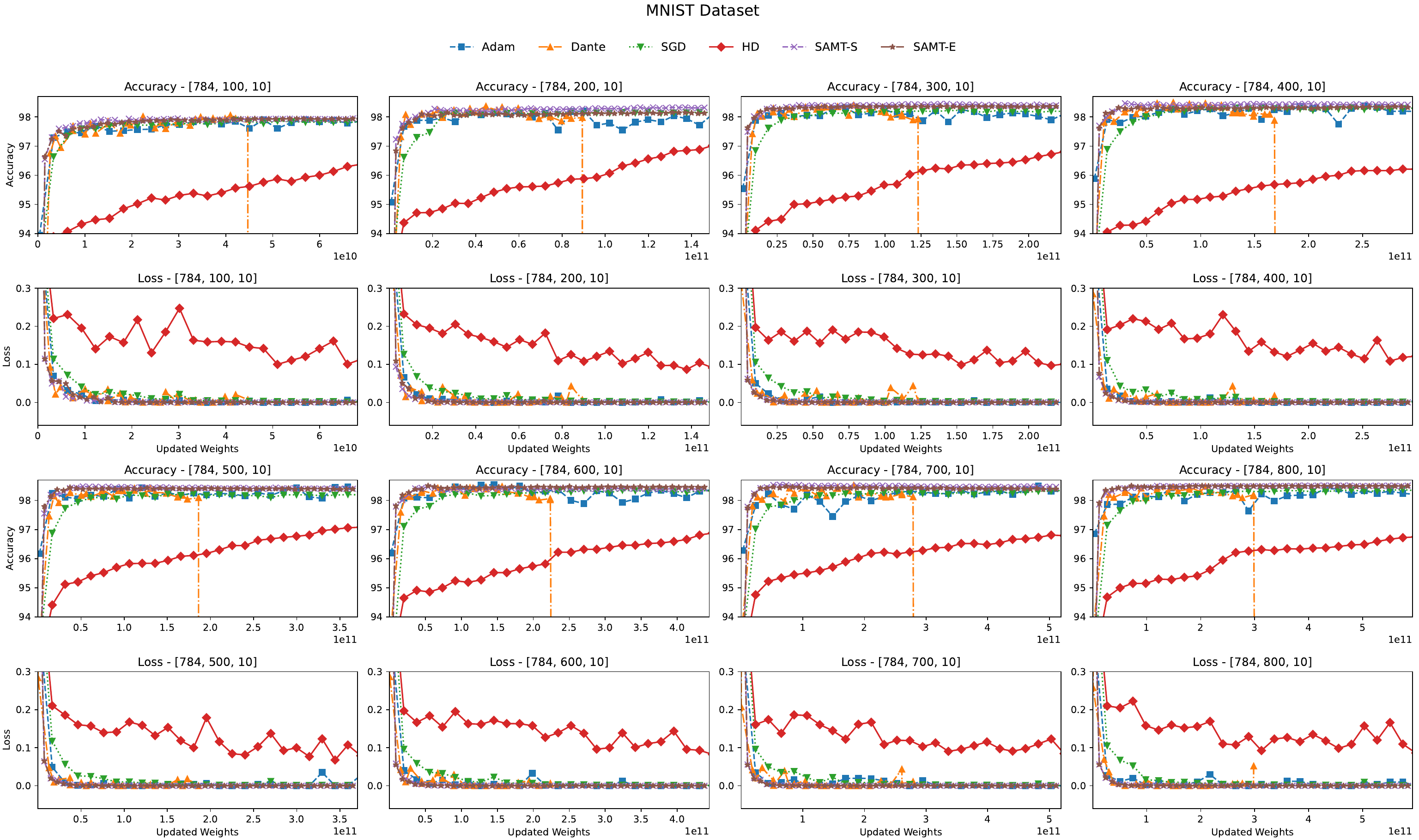}
    \caption{Comparison of MLP performance for different methods: Test accuracy and loss curves on the MNIST dataset. } 

\label{fig:mlp-mnist}
\end{figure}

\begin{figure}
\centering
\includegraphics[width=1.0\linewidth]{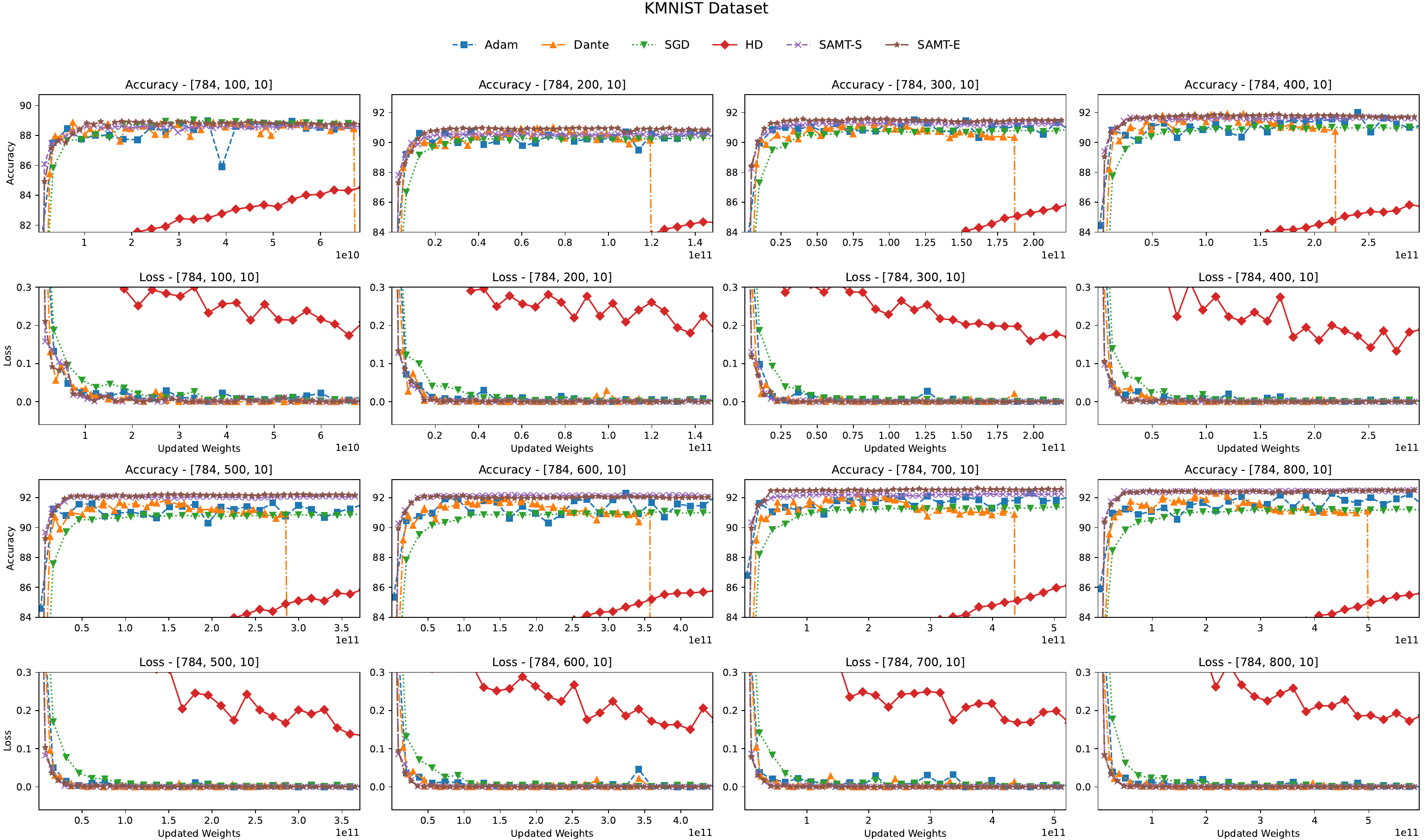}
    \caption{Comparison of MLP performance for different methods: Test accuracy and loss curves on the KMNIST dataset. } 

\label{fig:mlp-kmnist}
\end{figure}

Figures \ref{fig:mlp-mnist} and \ref{fig:mlp-kmnist} illustrate the comparative results of MLP-based experiments conducted on two distinct datasets. From the experimental curves, it is evident that while the Dante algorithm exhibits strong performance in the early stages of training, it experiences a notable decline in performance during the middle and later phases. The Adam optimizer, on the other hand, demonstrates fluctuations in performance toward the later stages of training. However, the HD algorithm demonstrated relatively slow convergence across both datasets and was unable to attain competitive accuracy. Additionally, although the fixed step-size SGD algorithm converges more slowly in the early phase, it yields more stable results in the later stages. In comparison, the proposed algorithm outperforms the above methods for most network dimensionalities, achieving both faster convergence and higher test accuracy.

In addition to the conventional handwriting recognition dataset, we further conducted experimental comparisons on the Fashion-MNIST dataset, as summarized in Table \ref{tab:fmnist}. Overall, the Adam optimizer exhibits certain advantages in low architecture dimensions. However, as the network dimensionality increases, our proposed scalar-version algorithm (SAMT-S) demonstrates more stable performance and achieves competitive accuracy for most dimensions. Notably, in higher-dimensional settings, the element-version (SAMT-E) of our algorithm exhibits superior performance.

Based on the experimental results for the three MNIST datasets, the above chart and table indicate that as the dimensionality of the hidden layers increases, our proposed algorithm demonstrates notable advantages over existing methods. While its performance may be slightly inferior or comparable to baselines in lower-dimensional settings, it exhibits greater robustness in higher-dimensional conditions, largely attributed to the adaptive adjustment of trainable step size via a meta-learning strategy. This observation is further supported by the results of the subsequent ablation study in Section \ref{sec:abstudy}. For a more comprehensive analysis of our proposed method, the comparative experiments with the AM algorithms based on full-batch samples and gradient will be presented in Section \ref{sec:addition}.

\begin{table}[h]
\centering
\caption{MLP experiment results on FMNIST dataset: Test accuracy (\%). Bold indicates the best performance, and underlined values denote the second-best in each column.}
\label{tab:fmnist}
\resizebox{\textwidth}{!}{
\begin{tabular}{c|cccccccc}
\hline
\multirow{2}{*}{Methods} & \multicolumn{8}{c}{Architecture Dimension}  \\ 
\cline{2-9}
~ & 100 & 200 & 300 & 400 & 500 & 600 & 700 & 800 \\
\hline
Dante & 88.84 & 89.48 & \underline{89.82} & 89.74 & 90.04 & 89.97 & 90.04 & 90.23 \\
\hline
HD & 86.86 & 87.32 & 87.43 & 87.59 & 87.61 & 87.55 & 87.72 & 87.68 \\
\hline
SGD & \underline{88.94} & 89.54 & 89.60 & \underline{89.98} & \underline{90.10} & \underline{90.06} & \underline{90.17} & \textbf{90.38} \\
\hline
Adam & \textbf{88.98} & \textbf{90.04} & 89.76 & 89.79 & 89.97 & 89.96 & 90.04 & 90.03 \\
\hline
SAMT-S (our) & 88.76 & \underline{89.69} & \textbf{89.83} & \textbf{90.14} & \textbf{90.23} & \textbf{90.17} & 90.15 & 90.00 \\
\hline
SAMT-E (our) & 88.84 & 89.30 & 89.59 & 89.95 & 90.01 & 90.01 & \textbf{90.27} & \underline{90.35} \\
\hline
\end{tabular}
}
\end{table}

\subsubsection{CNN Experiments}

We further conducted CNN experiments on two benchmark datasets: CIFAR-10 and CIFAR-100. In contrast to the previous experiments based on MLP, the CNN architecture incorporates convolutional layers, max pooling operations, and fully connected layers, with nonlinear activation functions applied between layers to produce the final network output. The CNN architecture is organized as follows: 
\begin{equation*}
    \text{Conv–Activation–MaxPool–Conv–Activation–MaxPool–Dropout–Fully Connected}.
\end{equation*}

\begin{figure}[H]
\centering
\includegraphics[width=0.9 \linewidth]{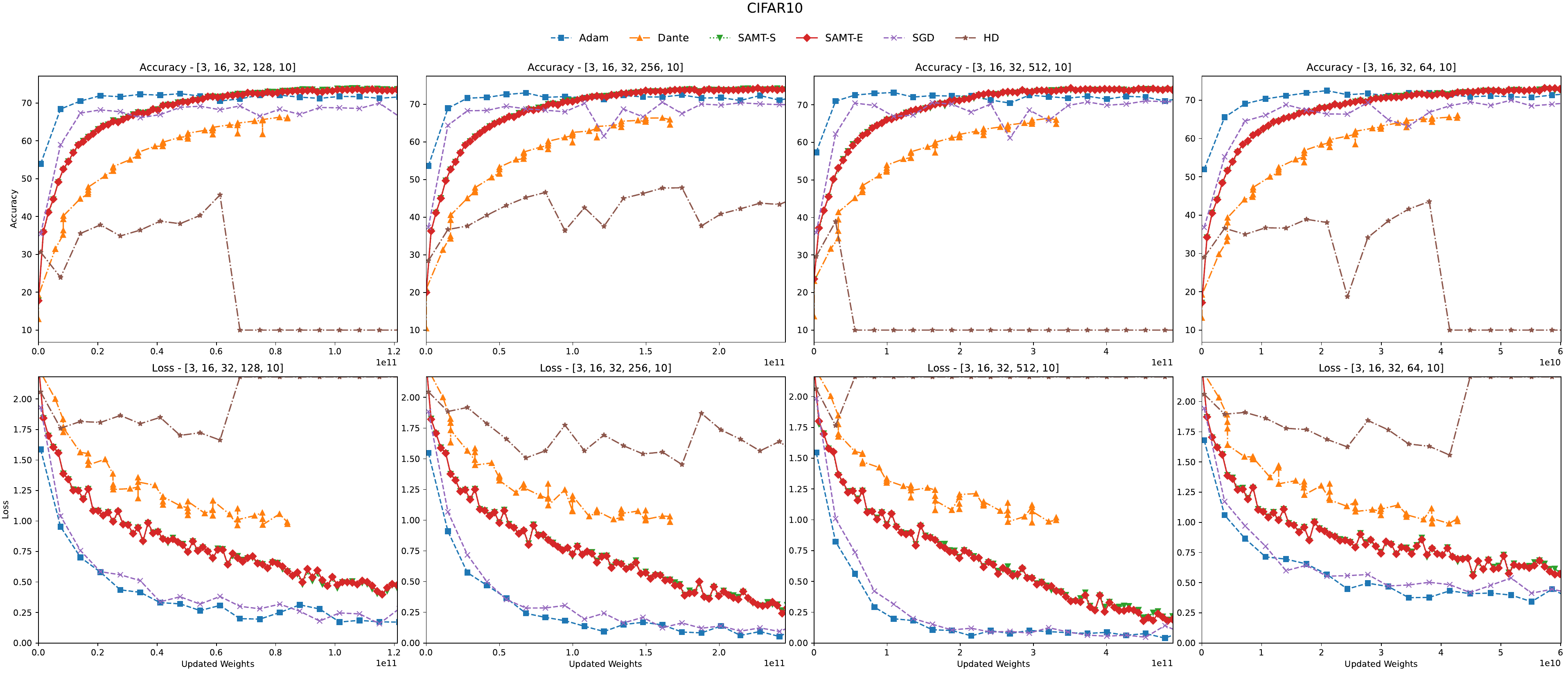}
    \caption{Comparison of CNN performance for different methods: Test accuracy and loss curves for CIFAR10 dataset. } 

\label{fig:cifar10}
\end{figure}

Figure \ref{fig:cifar10} presents the experimental results on the CIFAR-10 dataset using network structures of varying dimensionalities. The performance of the HD algorithm is slightly inferior on both datasets, with noticeable convergence issues in the later stages of training. The Adam optimizer exhibits rapid convergence in the early training phase and stabilizes in the later phase, benefiting from the momentum term and adaptive learning rate scaling based on gradient magnitudes. In contrast, the SGD optimizer with a constant step size also converges quickly at the beginning but suffers from accuracy fluctuations in subsequent epochs. The Dante algorithm, which adopts an alternating optimization strategy, demonstrates relatively poor convergence behavior overall. In comparison, the proposed algorithm converges more slowly during the initial stages but achieves greater stability in the later stages and ultimately outperforms the aforementioned methods in terms of final test performance.

\begin{figure}[H]
\centering
\includegraphics[width=0.9 \linewidth]{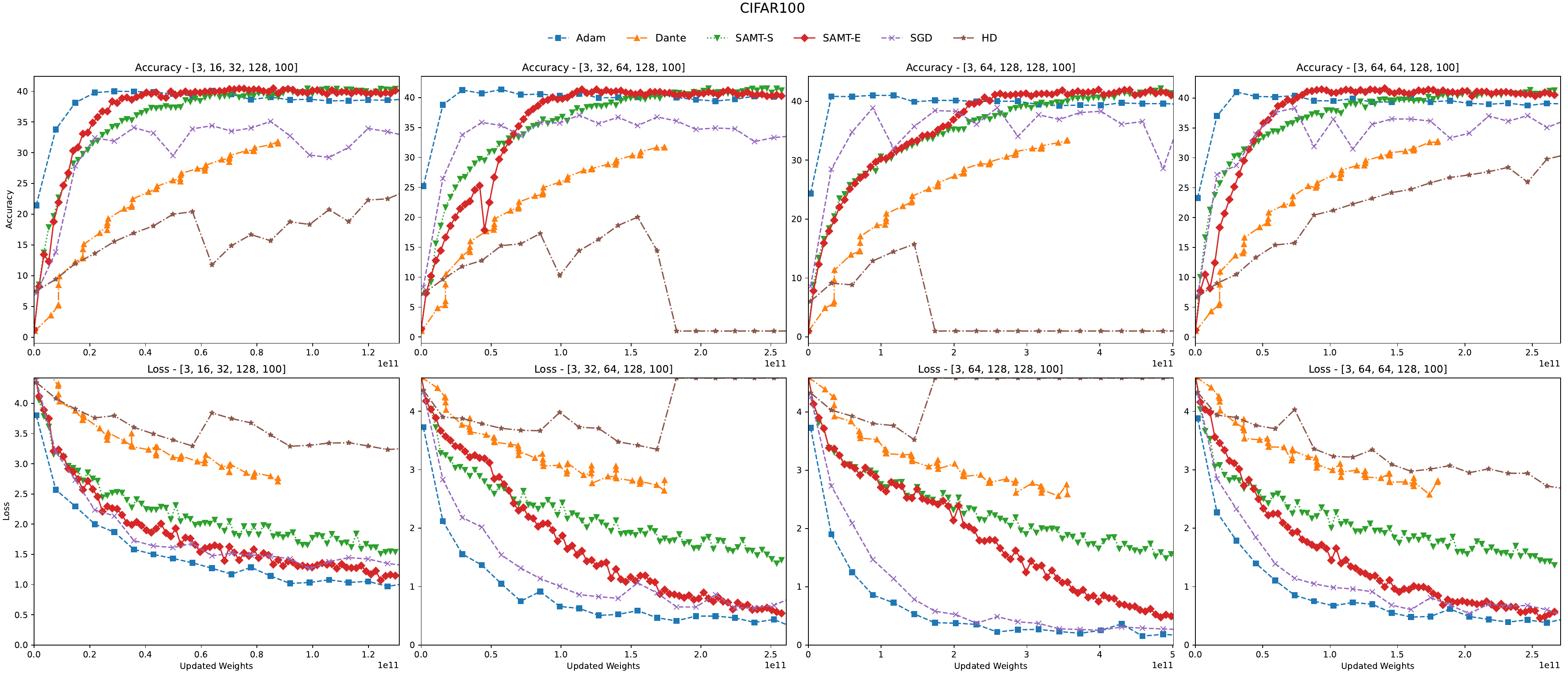}
    \caption{Comparison of CNN performance for different methods: Test accuracy and loss curves for CIFAR100 dataset. } 

\label{fig:cifar100}
\end{figure}

Similarly, Figure \ref{fig:cifar100} shows the comparative experimental results on the CIFAR-100 dataset. As shown in the figure, the scalar version of the proposed algorithm performs slightly worse than the non-scalar variant during the early training phase; however, in the later stages, its performance tends to surpass that of the non-scalar version. The performance trajectories of the other baseline algorithms are largely consistent with those observed on the CIFAR-10 dataset. Overall, in the context of CNN-based experiments, the proposed algorithm demonstrates clear advantages for the two datasets.

In summary, the proposed SAMT method demonstrates better performance for multiple datasets and various neural network architectures. Although it may exhibit slightly inferior results compared to other algorithms on low-dimensional networks, in high-dimensional settings, the proposed method can autonomously learn more suitable step sizes through meta-learning strategies within an alternating optimization framework, leading to more stable iterative updates. In the following sections, we analyze the parameter sensitivity of SAMT and present related ablation studies.

\subsection{Sensitivity Analysis}
\begin{figure}
\centering
\includegraphics[width=1.0\linewidth]{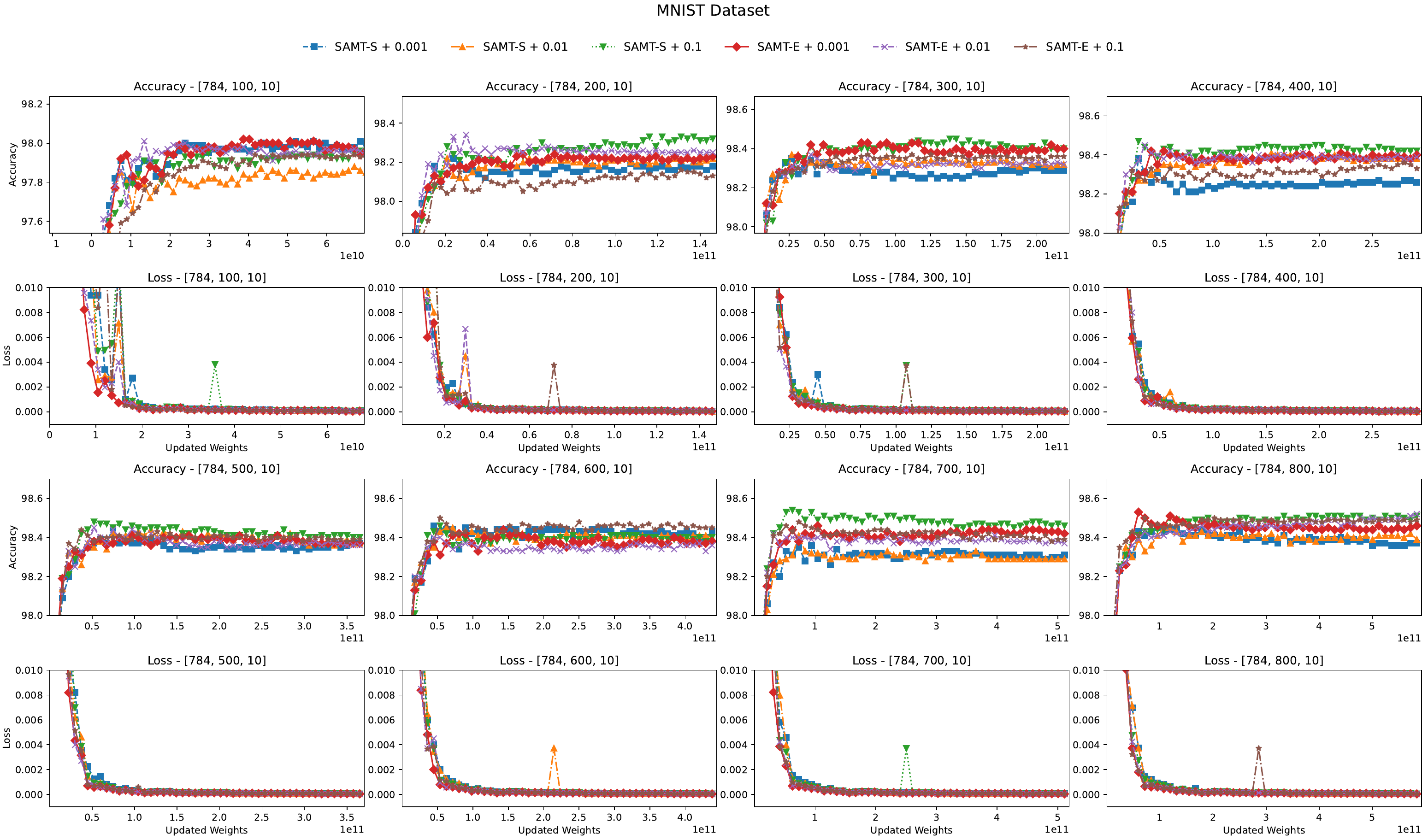}
    \caption{Different $\bar{\eta}^{0}$: Test accuracy and loss curves for MNIST dataset. } 

\label{fig:mlp-mnist-diffeta}
\end{figure}

To further assess the robustness of our proposed algorithm with respect to the initial step size $\bar{\eta}^{0}$, as shown in Figure \ref{fig:mlp-mnist-diffeta}, we conduct a sensitivity analysis in this section. By initializing the step size with different values $\bar{\eta}^{0} \in \{0.1,0.01,0.001\}$, we systematically evaluate the sensitivity of various comparative algorithms to changes in the initial step size $\bar{\eta}^{0}$. Owing to space restrictions, the figure of the KMNIST dataset of the proposed method is provided in Appendix \hyperref[sec:B.1]{B.1}.

Taking the MNIST dataset as an illustrative example, when the initial step size is gradually decay, it is observed that the test accuracy exhibits noticeable fluctuations in low-dimensional hidden layers, though the variation remains within a controllable range (±0.2). As the hidden layer dimensionality increases, the scalar version of the SAMT algorithm shows reduced stability in certain cases—particularly when the hidden dimension is set to 700. Nevertheless, under most other high-dimensional settings, the influence of initial step size variations on algorithm performance is minimal, and the resulting accuracy curves remain consistently stable.

In subsequent experiments about Section \ref{sec:abstudy} and \ref{sec:projectionstyle}, to facilitate experimental design, we fixed $\bar{\eta}^{0}=0.1$ and conducted multiple MLP experiments comparisons based on the MNIST series dataset. On the one hand, we designed an ablation study to investigate the interrelationships among the learnable step size of the eta model $\psi$. On the other hand, to examine how different projection functions influence the trainable step size, we analyzed the performance of the proposed algorithm with two different projection strategies. Detailed results and discussions are provided in the subsequent two subsections.

\begin{figure}
\centering
\includegraphics[width=1.0 \linewidth]{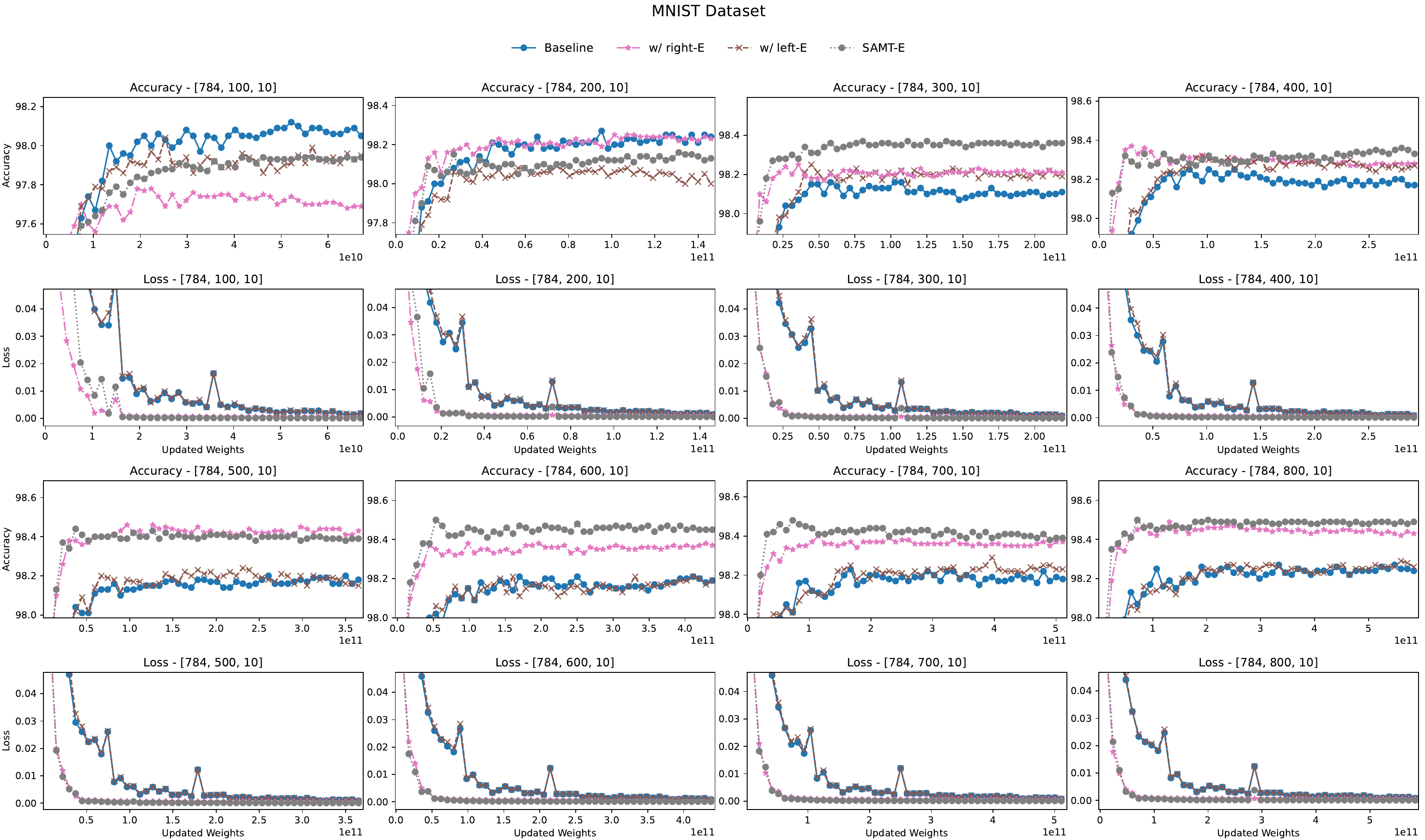}
    \caption{Ablation experiments for SAMT-E method: Test accuracy and loss curves for MNIST dataset. } 

\label{fig:mlp-mnist-abstudy-e}
\end{figure}

\subsection{Ablation Studies}
\label{sec:abstudy}
In addition to evaluating the algorithm's overall performance for different datasets, it is also essential to examine its performance under varying settings. To this end, we conducted a series of ablation experiments to assess the algorithm’s effectiveness under different parameter combinations. 

For the sake of analytical clarity, in this part, we discuss the ablation studies on the element version of \eqref{eq:update_etann}. Specifically, four experimental configurations are designed: \textbf{Baseline, left-E, right-E, and STAM-E}. The Baseline employs a fixed step size without any trainable parameters. The left-S and right-S correspond to the left and right components of the summation in \eqref{eq:update_etann}, respectively. STAM-E represents the complete model incorporating both components. Due to page constraints, the figure about ablation performance of the scalar version of the proposed method is provided in Appendix \hyperref[sec:B.2]{B.2}.

As shown in Figure \ref{fig:mlp-mnist-abstudy-e}, compared to the baseline model, the performance of the right-E configuration is consistently superior to that of left-E for both MNIST series datasets, underscoring the critical role of the learnable step size in the overall training process. Notably, on these dataset, right-E even outperforms the full STAM-E method in certain dimensions, further highlighting its effectiveness. On the MNIST dataset, the complete STAM-E model—which integrates both left-E and right-E through a weighted summation—demonstrates the complementary benefits of combining a scaling factor with a learnable step size. While the left-E component alone yields only marginal improvements over the baseline, its integration with right-E enables the full algorithm to achieve the highest test performance, indicating a synergistic interaction between the two components.

It is worth noting that the ablation experiments demonstrate that, compared to the AM algorithm with a fixed step-size strategy (baseline), the SAMT algorithm with trainable step sizes exhibits significant advantages across more dimensions. This further highlights the necessity of incorporating trainable step sizes within our proposed alternating optimization framework. By iteratively updating the trainable step sizes during the alternating optimization process, the model training can be more effectively optimized.

\subsection{Different Projection Style: Tanh vs Sigmoid}
\label{sec:projectionstyle}
\begin{figure}[H]
\centering
\includegraphics[width=1.0\linewidth]{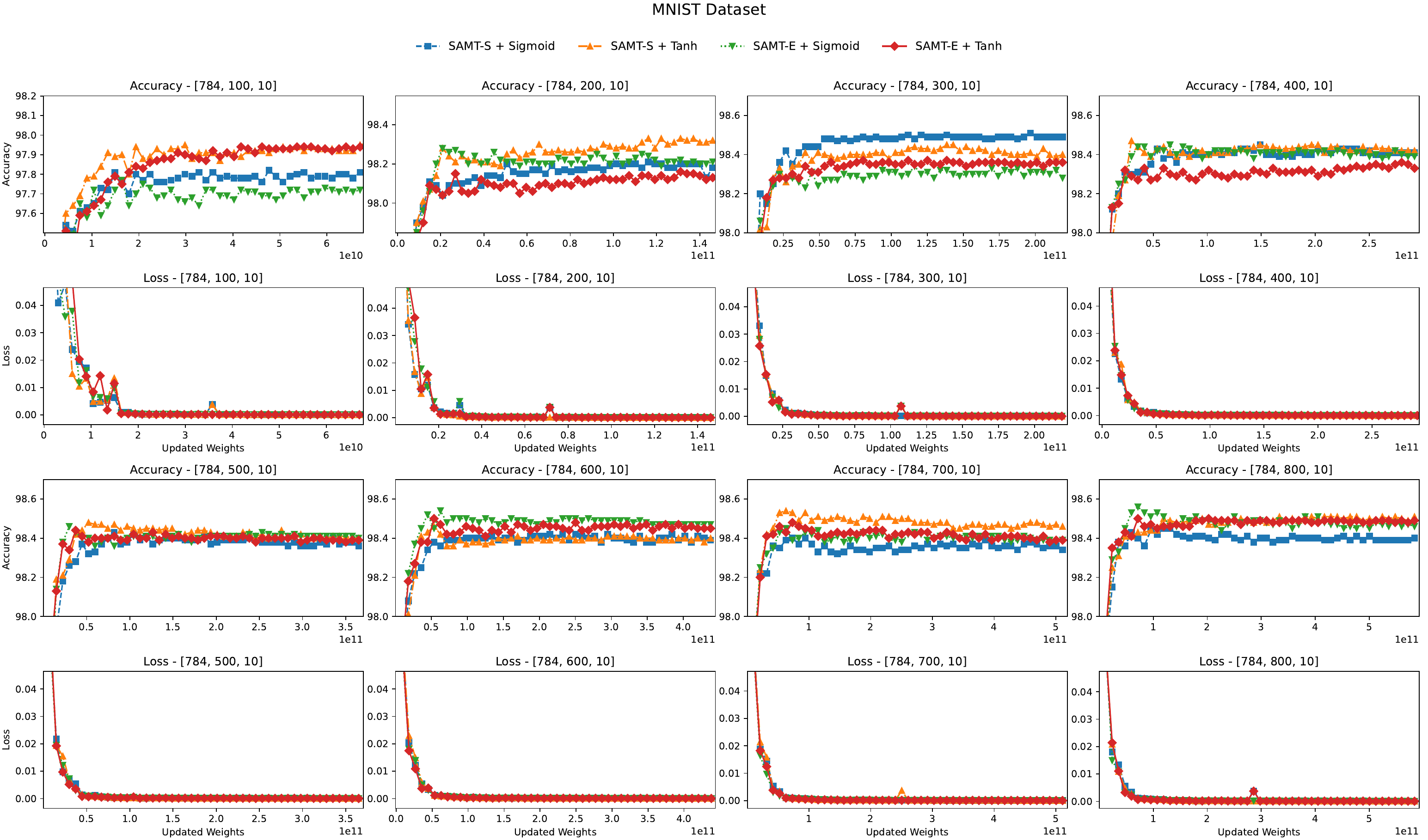}
    \caption{ Different projection style (Tanh vs Sigmoid): Test accuracy and loss curves for MNIST dataset. } 

\label{fig:mlp-mnistvs}
\end{figure}

Specially, Due to the learned parameters $\hat{\eta}^{t}$ in \eqref{eq:update_etann} are required to be projected to the range [0,1], to explore the influence of different projection functions on the output of the eta model $\psi$, we evaluated the effect of different projection functions, e.g., Tanh ($0.5 \cdot (tanh(x) + 1)$) and Sigmoid, by substituting them within the proposed method. Indeed, compared to sigmoid, tanh provides a broader dynamic range, facilitating more flexible control over trainable step sizes during optimization. We using the MNIST and KMNIST datasets as benchmarks; the corresponding test performance is presented in Figure \ref{fig:mlp-mnistvs}. For brevity, the figure about the projection performance of the scalar version and the KMNIST dataset of the proposed method is provided in Appendix \hyperref[sec:B.3]{B.3}.

From the comparative analysis of the two projection functions, it is observed that on the MNIST dataset, although the sigmoid projection exhibits advantages in specific dimensions such as [784, 300, 10] and [784, 600, 10], the Tanh-based projection function demonstrates superior performance across a broader range of dimensions. A similar trend is observed on the KMNIST dataset, where the sigmoid function shows localized benefits, most notably in the [784, 800, 10] dimension, yet the tanh projection yields better performance across the remaining dimensions.

Further examination of the output distributions reveals that, while both functions produce outputs constrained to the interval (0,1), the sigmoid function tends to concentrate its output values near the boundaries of this interval. This behavior increases the likelihood of generating extreme values, potentially leading to saturation effects during optimization. In contrast, the tanh-based projection produces outputs that are more centered within the interval, resulting in a smoother and more evenly distributed step size. This characteristic can mitigate over-saturation in step size adjustment.

\section{Additional Empirical Investigations}
\label{sec:addition}

Aiming to provide a more comprehensive evaluation of our proposed algorithm for different scenarios, we designed three supplementary experiments: (1) a comparative study of full-batch-based alternating optimization methods; (2) an evaluation of performance on regression tasks; (3) the type extensibility of trainable step sizes of the SAMT method.

\subsection{AM-based Algorithm Comparison Experiment}
In this section, to support a clearer analytical comparison, we present a comparative analysis between our proposed SAMT-E method and several state-of-the-art AM algorithms. The evaluation is conducted across different hidden layer dimensions to assess the performance of each approach.

\begin{table}[h]
\centering
\caption{Comparison of MLP experiments for SAMT-E and AM (Full-Batch Gradient) algorithms on MNIST series datasets: Test Accuracy (\%).}
\label{tab:AM-mnist-all}
\resizebox{\textwidth}{!}{
\begin{tabular}{c|cccccccc}
\hline
~ & \multicolumn{8}{c}{MNIST Dataset}         \\
\hline
\multirow{2}{*}{Methods} & \multicolumn{8}{c}{Architecture Dimension}  \\ 
\cline{2-9}
~ & 100 & 200 & 300 & 400 & 500 & 600 & 700 & 800 \\
\hline
dlADMM & 80.22 & 88.62 & 91.80 & 92.90 & 92.97 & 93.96 & 94.19 & 95.12 \\
\hline
mDLAM &  82.73 & 89.22 & 91.92 & 92.33 & 91.82 & 89.07 & 88.68 & 85.16 \\
\hline
SAMT-E &  97.95 & 98.17 & 98.39 & 98.39 & 98.46 & 98.52 & 98.48 & 98.51 \\
\hline
~ & \multicolumn{8}{c}{KMNIST Dataset}         \\
\hline
\multirow{2}{*}{Methods} & \multicolumn{8}{c}{Architecture Dimension}  \\ 
\cline{2-9}
~ & 100 & 200 & 300 & 400 & 500 & 600 & 700 & 800 \\
\hline
dlADMM & 51.32 & 61.66 & 67.86 & 70.60 & 72.85 & 73.74 & 74.98 &  76.33 \\
\hline
mDLAM &  54.93 & 64.75 & 70.39 &  66.36 & 64.33 & 45.63 & 44.21 & 35.33 \\
\hline
SAMT-E &  88.98 & 91.05 & 91.61 & 91.92 & 92.25 & 92.15 & 92.64 & 92.52 \\
\hline
~ & \multicolumn{8}{c}{FMNIST Dataset}         \\
\hline
\multirow{2}{*}{Methods} & \multicolumn{8}{c}{Architecture Dimension}  \\ 
\cline{2-9}
~ & 100 & 200 & 300 & 400 & 500 & 600 & 700 & 800 \\
\hline
dlADMM &  76.06 & 80.66 & 82.77 &  82.88 & 83.74 & 84.43 & 84.12 & 84.19 \\
\hline
mDLAM &  75.81 & 80.41 &  77.83 & 50.11 & 10.83 & 8.53 & 13.31 & 12.30 \\
\hline
SAMT-E & 88.84 & 89.30 & 89.59 & 89.95 & 90.01 & 90.01 & 90.27  & 90.35 \\
\hline
\end{tabular}
}
\end{table}

Table \ref{tab:AM-mnist-all} represents the performance of our proposed algorithm in comparison with dlADMM~\cite{wang2019admm} and mDLAM~\cite{wang2022accelerated}, two representative methods employing full-batch alternating optimization. The experimental results clearly indicate that the randomized alternating optimization approach achieves notably faster convergence and higher accuracy, highlighting its superiority over full-batch counterparts in MLP training tasks. 

\subsection{Regression Task}

Table \ref{tab:regexp} presents the performance of various algorithms on regression tasks. There are three regression datasets in this part\footnote{\url{https://archive.ics.uci.edu/datasets}}. Lower values indicate lower testing errors and better regression fitting performance. Overall, the SAMT-E algorithm achieves better average performance compared to the other three algorithms. In particular, for the Bike dataset, the SAMT-S algorithm demonstrates better test error. Interestingly, as for the remaining two regression datasets, compared with the SAMT-S algorithm, the SAMT-E exhibits more favorable performance. Furthermore, the Adam algorithm exhibits notable advantages in regression tasks.

\begin{table*}[thb]\centering
    \caption{Comparison of test error (MSE) on regression datasets. Bold indicates the best performance, and underlined values denote the second-best in each column.}
    \label{tab:regexp}
    \resizebox{0.8\textwidth}{!}{
    \begin{tabular}{c c c c c}
        \toprule
        Methods & Bike & Pol & Kegg & Average \\
        \midrule
        Adam & 0.1736 & \underline{0.1034} & \underline{0.0320} & \underline{0.1030} \\
        SGD & 0.1732 & 0.1105 & 0.3506 & 0.2114 \\
        HD &  \underline{0.1731} & 0.1184 & 0.0377 & 0.1097 \\
        Dante & 0.2350 & 0.1360 & 0.0357 & 0.1356 \\
        SAMT-S (our) & \textbf{0.1728} & 0.1155 & 0.0365 & 0.1083 \\
        SAMT-E (our) & 0.1734 & \textbf{0.0995} & \textbf{0.0315} & \textbf{0.1015} \\
        \bottomrule
    \end{tabular}
    }
\end{table*}

\subsection{Type Extensibility of Trainable Step Sizes for SAMT method}

As discussed in Section \ref{sec:onlinestepsize}, we now further explore different types of trainable step sizes. Constrained by the length of this article, we primarily focus on two other representative types: row-wise (SAMT-R) and column-wise (SAMT-C) step sizes by leveraging the dimension-broadcast mechanisms commonly available in deep learning frameworks~\cite{paszke2017automatic}. It is worth noting that the proposed SAMT method can be readily extended to accommodate any desired step size type with minimal modification.

Taking the MLP experimental as an example, the row-wise configuration specifies that all elements within the same row of the weight matrix share a trainable stride parameter, whereas the column-wise configuration enforces parameter sharing across each column. All parameter settings in this section are kept consistent with those used in the previous MLP experiments. For clarity of analysis, we present only the impact of different step size strategies on test accuracy.

\begin{figure}[H]
\centering
\includegraphics[width=1.0\linewidth]{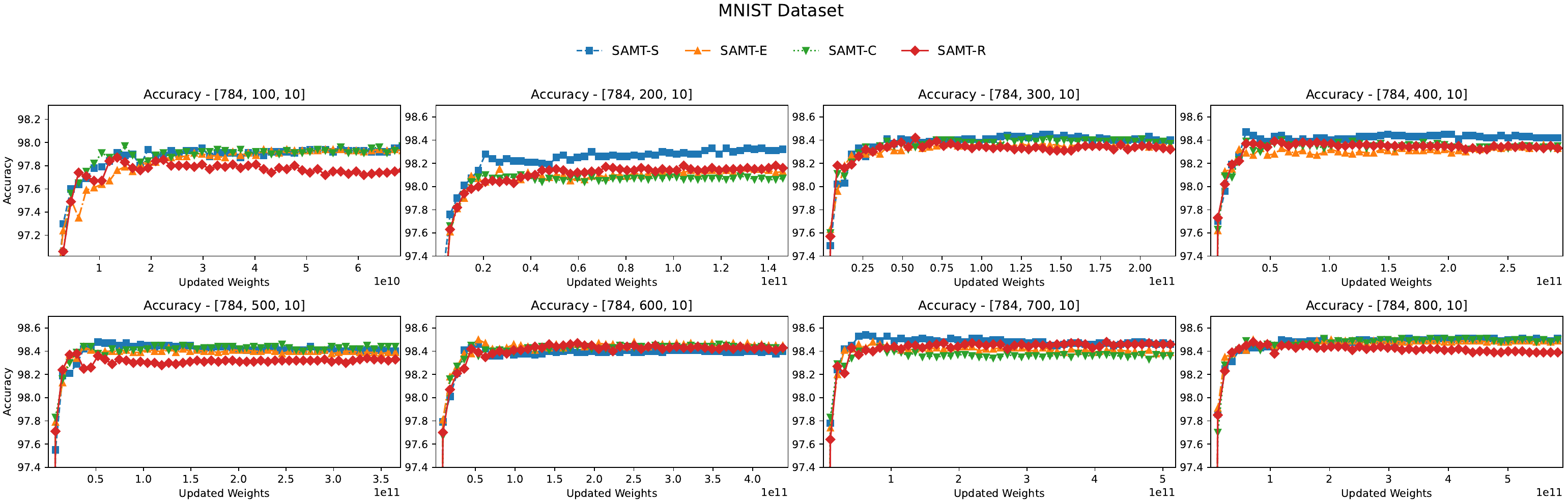}
    \caption{ Test accuracy under different step size types on the MNIST dataset. } 

\label{fig:mlp-mnist-othertype}
\end{figure}

\begin{figure}[H]
\centering
\includegraphics[width=1.0\linewidth]{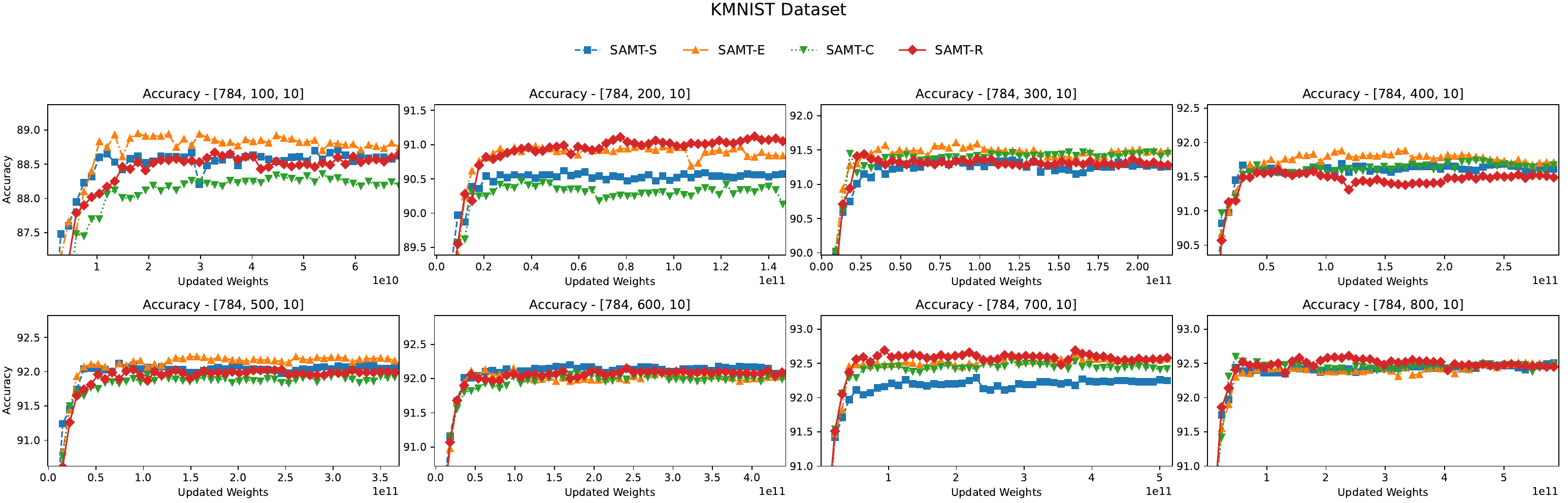}
    \caption{ Test accuracy under different step size types on the KMNIST dataset. } 

\label{fig:mlp-kmnist-othertype}
\end{figure}

\begin{figure}[H]
\centering
\includegraphics[width=1.0\linewidth]{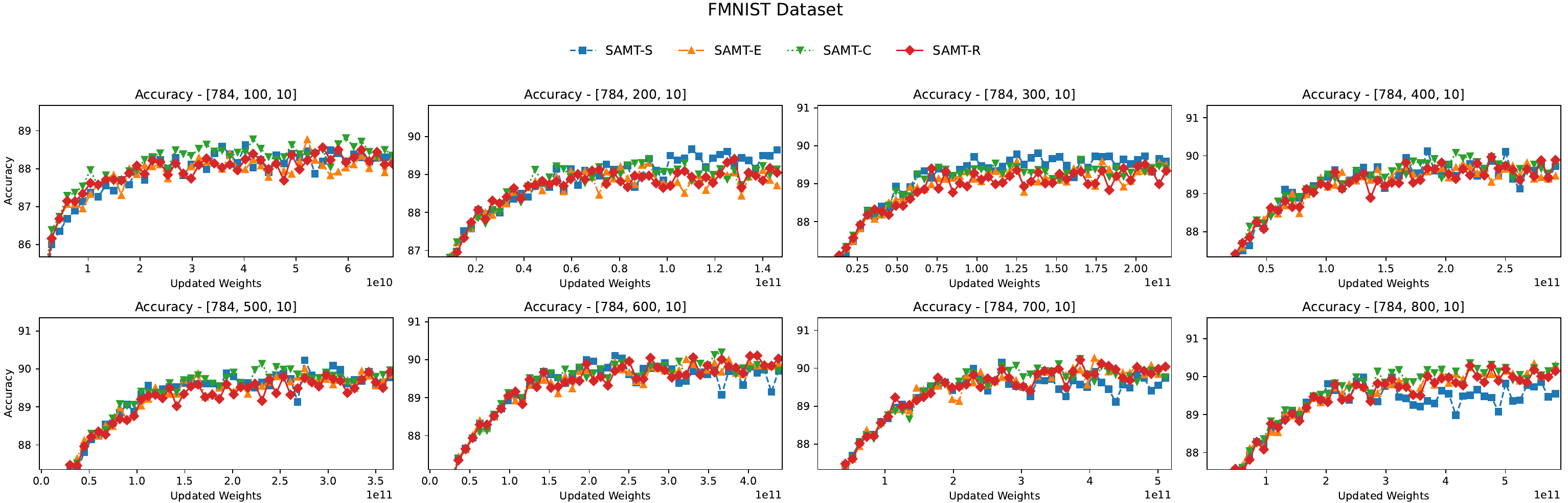}
    \caption{ Test accuracy under different step size types on the FMNIST dataset. } 

\label{fig:mlp-fmnist-othertype}
\end{figure}

As shown in Figure \ref{fig:mlp-mnist-othertype}, on the MNIST dataset, the SAMT-R accuracy curve demonstrates the lowest performance in most dimension configurations. Although SAMT-C exhibits trends similar to SAMT-R as the dimensionality changes, it occasionally outperforms other variants in certain configurations, such as [784, 100, 10]. In Figure \ref{fig:mlp-kmnist-othertype} and \ref{fig:mlp-fmnist-othertype}, for the KMNIST dataset, in contrast, the performance of SAMT-R is more favorable compared to SAMT-C, and for the FMNIST dataset, the SAMT-C method achieves slightly better performance compared to SAMT-R. Overall, the results indicate that different step size types are suitable for different types of datasets, underscoring the significance of exploring various step size strategies for neural network training.

\section{Limitations and Future work}
\label{sec:future}
It should be acknowledged that while the proposed algorithm has certain advantages in solving neural network problems, it still exhibits limitations in some aspects:
\begin{itemize}
    \item Due to the requirement for online learning of trainable step sizes, maintaining the same computational graph structure for the original model. This constraint may significantly slow down the training process when dealing with deep neural networks with numerous layers, e.g., ResNet and VGG neural networks.
    
    \item In our theoretical analysis, we established the convergence guarantee of the proposed algorithm under mild conditions. However, a formal proof of its convergence property under stronger assumptions remains an open problem.
\end{itemize}

In our future work, we will focus on optimizing the trainable step size implementation mentioned above to train deeper neural networks. Furthermore, our algorithm is inherently flexible and can be readily extended to group multiple layers into a single block, enabling block-wise alternating updates under various configurations. The specific mathematical formulation of the grouped three-layer neural network is given as follows:
\begin{equation}
\begin{aligned}
    \label{process:am_multiblock}
    &W_{1,2}^{t} \leftarrow \arg\min_{W_{1,2}}\mathbb{E}_{(\mathbf{x},\mathbf{y})\sim\mathcal{S}}  \| \phi_{W_{1,2}}( \mathbf{x};  \cdot ) - \mathbf{y} \|^{2}  , \nonumber  \\ 
    &W_3^{t} \leftarrow \arg\min_{W_3}\mathbb{E}_{(\mathbf{x},\mathbf{y})\sim\mathcal{S}}  \| \phi_{W_3}( \mathbf{x};  \cdot ) - \mathbf{y} \|^{2}  ,  
\end{aligned}
\end{equation}
where $\phi_{W_{1,2}}( \mathbf{x}; \cdot ) = \mathcal{F}_3^{t-1} \circ \sigma_{2} \circ \mathcal{F}_{2} \circ \sigma_{1} \circ \mathcal{F}_{1}(\mathbf{x}) $, and $W_3$ and $W_2$ need to be updated in the optimization process simultaneously. Additionally, we will establish the convergence property of the stochastic AM algorithm for neural network training under stronger assumptions in future work.

\section{Conclusion}\label{conclusion}
\label{sec:Conclusion}
In this work, we proposed a novel neural network training method, SAMT, which is based on the concept of stochastic AM and employs trainable step sizes for parameter updates. Additionally, we have conducted a theoretical convergence analysis, demonstrating that the proposed algorithm satisfies the convergence guarantees under mild assumptions. Beyond its theoretical guarantees, our AM strategy offers high flexibility and extensibility. It can be easily further extended to the combination of different step size types or various groups of blocks for alternating updates. Extensive experiments have shown that SAMT exhibits notable advantages over existing algorithms.

\section*{Acknowledgment}
\noindent{\bf Funding:} This research is supported by the National Natural Science Foundation of China (NSFC) grants 92473208, 12401415, the Key Program of National Natural Science of China 12331011, the 111 Project (No. D23017),  the Natural Science Foundation of Hunan Province (No. 2025JJ60009), and the Postgraduate Scientific Research Innovation Project of Hunan Province (No. LXBZZ2024114). 

\noindent{\bf Data Availability:} Enquiries about data/code availability should be directed to the authors.

\noindent{\bf Competing interests:} The authors have no competing interests to declare that are relevant to the content of this paper.

\section*{Appendix}
\label{appendix_list}

\subsection*{A. Proof Theorem}
\label{proof:final}
\subsection*{A.1 Proof Preliminaries}

The proof of Theorem \ref{theorem:final} is structured in three parts: a review of the necessary preliminaries, followed by the first and second components of the main proof argument. Each part will be discussed in detail below.

\begin{definition}
\label{def:operation}
    [Generalized Gradient AM Operator] Given a step size $\bar{\eta}$, the generalized gradient AM operator $\mathcal{Q}_i(W_1, W_2, \dots, W_L)$ is defined as:
\[
\mathcal{Q}_i(W_1, W_2, \dots, W_L) := W_i + \bar{\eta} \nabla_{i} \mathcal{J}(W_1, W_2, \dots, W_L),
\]
where $i = 1,2,\cdots,L$, and $\nabla_{i} \mathcal{J}$ denotes the partial gradient of the function $\mathcal{J}$ with respect to the parameter $W_i$. Concerning the Definition \ref{def:operation}, we have the following theories.
\end{definition}

\begin{lemma}\cite{choromanska2019beyond}
\label{lemma:leq}
For any $d = 1,2,\dots,L$, the gradient operator $\mathcal{Q}_d(W_1^{*},\dots,W_{d-1}^{*},W_{d},W_{d+1}^{*},\dots,W_L^{*})$ under Assumption~\ref{def:strongconN} (strong concavity) and Assumption~\ref{def:smoothN} (smoothness) with the step size choice $0 < \bar{\eta} \leq \frac{2}{\mu_d + \lambda_d}$ is contractive, i.e.,
\begin{equation}
\| \mathcal{Q}_d(W_1^{*},\dots,W_{d-1}^{*},W_{d},W_{d+1}^{*},\dots,W_L^{*}) - W_d^{*}\| \leq \left(1 - \frac{2\bar{\eta}\mu_d\lambda_d}{\mu_d + \lambda_d}\right)\|W_d - W^{*}_d\|   ,
\label{eq:contractivityTN}
\end{equation}
for all $W_d \in B_2(r_d,W_d^{*})$.
\label{lem:contru2N}
\end{lemma}

\begin{theorem}
For any $d$ from $1$ to $L$, there exists some radius $r_d > 0$ and a triplet $(\gamma_d,\lambda_d,\mu_d)$ such that $0 \leq \gamma_d < \lambda_d \leq \mu_d$, suppose that the function $\mathcal{J}(W_1^{*},W_2^{*},\dots,W_{d-1}^{*},W_d,W_{d+1}^{*},\dots,W_{L-1}^{*},W_L^{*})$ is $\lambda_d$-strongly concave (Assumption~\ref{def:strongconN}) and $\mu_d$-smooth (Assumption~\ref{def:smoothN}), and that the GS ($\gamma_d$) condition of Assumption~\ref{def:GSN} holds. Then the generalized gradient AM operator $\mathcal{Q}_d(W_1,W_2,\dots,W_L)$ with step size $\bar{\eta}$ such that $0 < \bar{\eta} \leq \min_{i = 1,2,\dots,L}\frac{2}{\mu_i + \lambda_i}$ is contractive over a ball $B_2(r_d,W_d^{*})$, i.e.,
\begin{equation}
\|\mathcal{Q}_d(W_1,W_2,\dots,W_L) - W_d^{*}\| \leq (1-\xi\bar{\eta})\|W_d - W_d^{*}\| + \bar{\eta}\gamma\sum_{\substack{i=1 \\ i\neq d}}^L\|W_i-W_i^{*}\|    ,
\end{equation}
where $\gamma := \max_{i=1,2,\dots,L}\gamma_i$, and $\xi := \min_{i = 1,2,\dots,L}\frac{2\mu_i\lambda_i}{\mu_i + \lambda_i}$.
\label{thm:contractivityG1wideN}
\end{theorem}

\subsection*{A.2 Proof of Theorem~\ref{thm:contractivityG1wideN}}
\begin{equation*}
\begin{aligned}
    & \|\mathcal{Q}_d(W_1,W_2,\dots,W_L) - W_d^{*}\| \\
    =& \|W_d + \bar{\eta}\nabla_d \mathcal{J}(W_1,W_2\dots,W_L) - W_d^{*}\|     \\
    \leq& \|W_d + \bar{\eta}\nabla_d \mathcal{J}(W_1^{*},\dots,W_{d-1}^{*},W_d,W_{d+1}^{*},\dots,W_L^{*}) - W_d^{*}\|  + \bar{\eta}\|\nabla_d \mathcal{J}(W_1,\dots,W_d,\dots,W_L)     \\
    &\quad - \nabla_d \mathcal{J}(W_1^{*},\dots,W_{d-1}^{*},W_d,W_{d+1}^{*},\dots,W_L^{*})\|      \\
    \leq &  \left(1 - \frac{2\bar{\eta}\mu_d\lambda_d}{\mu_d + \lambda_d}\right)\|W_d - W_d^{*}\| + \bar{\eta} \gamma_d \sum_{\substack{i=1 \\ i\neq d}}^L\|W_i - W_i^{*}\|  \\
    \leq& (1-\xi\bar{\eta})\|W_d - W_d^{*}\| + \bar{\eta}\gamma\sum_{\substack{i=1 \\ i\neq d}}^L\|W_i-W_i^{*}\|   ,
\end{aligned}
\end{equation*}
where the first inequality follows from the triangle inequality, and the second inequality follows from the contractivity of $T$ from \eqref{eq:contractivityTN} from Lemma~\ref{lem:contru2N} and the GS condition. The last inequality follows from $\gamma := \max_{i=1,2,\dots,L}\gamma_i$, and $\xi := \min_{i = 1,2,\dots,L}\frac{2\mu_i\lambda_i}{\mu_i + \lambda_i}$.    \\

\subsection*{A.3 Proof of Theorem~\ref{theorem:final}}
\subsubsection*{Part I: Proof of Theorem~\ref{theorem:final} }
\label{proof:part1}
According to the update rule of $W$ in Algorithm \ref{algo:learnable_scalar}, let $W_d^{t+1} = \Pi_d({\tilde W}_d^{t+1})$, here $\nabla_d \mathcal{J}^1$ is the gradient computed with respect to a single data sample, where ${\tilde W}_d^{t+1}:= W^t_d + \bar{\eta}^{t}\nabla_d \mathcal{J}^1(W_1^{t+1}, W_2^{t+1},\dots,W_{d-1}^{t+1},W_d^t,W_{d+1}^{t},\dots,W_L^t)$ is the update vector prior to the projection onto a ball $B_2(\frac{r_d}{2},W_d^0)$. Let $\Delta^{t+1}_d := W_d^{t+1} - W_d^{*}$ and ${\tilde\Delta}_d^{t+1} := {\tilde W}_d^{t+1} - W_d^{*}$. Therefore, we have 
\begin{align*}
    &\|\Delta^{t+1}_d\|^2 - \|\Delta^t_d\|^2  \\
    \leq&  \|{\tilde\Delta}^{t+1}_d\|^2 - \|\Delta^t_d\|^2\\
    =& \|{\tilde W}^{t+1}_d - W_d^{*}\|^{2} - \|W_d^t - W_d^{*}\|^{2} \nonumber\\
    =& \left<{\tilde W}_d^{t+1}  -  W_d^t, {\tilde W}_d^{t+1}  +  W_d^t  -  2W_d^{*}\right>.
\label{eqn:tmp1N}
\end{align*}
where the first inequality follows from the projective nonexpansive conclusion with respect to $\Pi_d$ ( i.e., $\| \Pi_d(z) - x^* \| \leq \|  z - x^* \|$ ), and the first equality follows from the simple algebra. Then we have that ${\tilde W}_d^{t+1} - W_d^t = \bar{\eta}^{t} \nabla{\hat{W}}_d^t$, where $\nabla {\hat{W}}_d^t:= \nabla_d \mathcal{J}^1(W_1^{t+1}, W_2^{t+1},\dots, W_{d-1}^{t+1},W_d^t,W_{d+1}^{t},\dots, W_L^t)$. Hence, we transformed the above inequality to the following formula:
\begin{align*}
&\|\Delta^{t+1}_d\|^2 - \|\Delta^t_d\|^2    \\
\leq& \left<\bar{\eta}^{t} \nabla {\hat{W}}_d^t,\bar{\eta}^{t} \nabla{\hat{W}}_d^t + 2(W_d^t - W_d^{*})\right>     \\
=& (\bar{\eta}^{t})^2( \nabla {\hat{W}}_d^t)^{\top}{ \nabla \hat{W}}_d^t + 2\bar{\eta}^{t}({\nabla \hat{W}}_d^t)^{\top}(W_d^t - W_d^{*})     \\
=& (\bar{\eta}^{t})^2\|{\nabla \hat{W}}_d^t\|^2 + 2\bar{\eta}^{t}\left<{ \nabla \hat{W}}_d^t,\Delta^t_d\right>.
\end{align*}

To further analyze the convergence of our algorithm in expectation, by the definition of 
$\nabla {\hat{W}}_d^t$, it holds that $\mathbb{E}[\nabla{\hat{W}}_d^t] = \nabla {W}_d^t$. We let $ \nabla W_d^t := \nabla_d \mathcal{J}(W_1^{t+1},W_2^{t+1},\dots,W_{d-1}^{t+1},W_d^t,W_{d+1}^{t},\dots,W_L^t)$, then taking expectations on both sides of the inequality above yields:
\begin{eqnarray}
\mathbb{E}[\|\Delta^{t+1}_d\|^2] \leq \mathbb{E}[\|\Delta^t_d\|^2] + (\bar{\eta}^{t})^2\mathbb{E}[\|{ \nabla \hat{W}}_d^t\|^2]+ 2\bar{\eta}^{t}\mathbb{E}[\left< \nabla {W}_d^t,\Delta^t_d\right>]   .
\label{eq:edeltauN}
\end{eqnarray}

By the self-consistency condition ( i.e., $W_d^{*} = \arg\max_{W_d\in\Omega_d} \mathcal{J}(W_1^{*},\dots,W_{d-1}^{*},W_d,W_{d+1}^{*},\dots,W_L^{*})$, and convexity of $\Omega_d$ . Let $\nabla {W}_d^{*} := \nabla_d \mathcal{J}(W_1^{*},W_2^{*},\dots,W_L^{*})$, we obtain
\[\left< \nabla W_d^{*},\Delta^t_d\right> = \left<\nabla_d \mathcal{J}(W_1^{*},W_2^{*},\dots,W_L^{*}),\Delta^t_d\right> \leq 0   ,
\]
combining this inequality with \eqref{eq:edeltauN}, and by the definition of $\nabla W^*_d$, so we can transform \eqref{eq:edeltauN} to below formula 
\begin{equation}
\begin{aligned}
\label{eq:mid}
&\mathbb{E}[\|\Delta^{t+1}_d\|^2]  \\
\leq&  \mathbb{E}[\|\Delta^t_d\|^2] + (\bar{\eta}^{t})^2\mathbb{E}[\|{ \nabla \hat{W}}_d^t\|^2]+ 2\bar{\eta}^{t}\mathbb{E}[\left< \nabla {W}_d^t,\Delta^t_d\right>]      \\
\leq& \mathbb{E}[\|\Delta^t_d\|^2] + (\bar{\eta}^{t})^2\mathbb{E}[\| \nabla {\hat{W}}_d^t\|^2]+ 2\bar{\eta}^{t} \mathbb{E}[\left< \nabla {W}_d^t - \nabla {W}_d^{*},\Delta^t_d\right>] .     
\end{aligned}    
\end{equation}

According the definition of \eqref{eq:mid}, we first process the third term $2\bar{\eta}^{t} \mathbb{E}[\left< \nabla {W}_d^t - \nabla {W}_d^{*},\Delta^t_d\right>]$ in \eqref{eq:mid}, therefore, for the next proof, by the Definition \ref{def:operation}, we define $\mathcal{Q}^t_d:= W_d^t + \bar{\eta}^{t} \nabla{W}^t_d$ and $\mathcal{Q}_d^{t*}:= W_d^{*} + \bar{\eta}^{t} \nabla {W}_d^{*}$. Concerning the third term in \eqref{eq:mid}, we can derive that
\small{
\begin{equation}
\label{eq:mid2}
\begin{aligned}
&\bar{\eta}^{t}\left< \nabla W_d^t -  \nabla W_d^{*},\Delta^t_d\right>\\
=& \left<\mathcal{Q}_d^t - \mathcal{Q}_d^{t*} - ( W^t_d - W_d^{*}),  W^t_d - W_d^{*}\right>\\
=& \left<\mathcal{Q}_d^t - \mathcal{Q}_d^{t*}, W^t_d - W_d^{*}\right> - \| W^t_d - W_d^{*}\|^2\\
=& \left<\mathcal{Q}_d^t - W_d^{*}, W^t_d - W_d^{*}\right>  - \| W^t_d - W_d^{*}\|^2  \\
\leq& \left\{(1-\bar{\eta}^{t}\xi)\|W_d^t-W_d^{*}\| + \bar{\eta}^{t}\gamma\left(\sum_{i=1}^{d-1}\|W_i^{t+1}-W_i^{*}\|\right.\right.\left.\left.+    \sum_{i=d+1}^{L}\|W_i^{t}-W_i^{*}\|\right)\right\}\| W^t_d - W_d^{*}\|  - \| W^t_d - W_d^{*}\|^2   \\
\leq&\left\{ (1 - \bar{\eta}^{t}\xi)\|\Delta_d^t\|  +  \bar{\eta}^{t}\gamma \left(\sum_{i=1}^{d-1}\|\Delta_i^{t+1}\|  +     \sum_{i=d+1}^{L}\|\Delta_i^t\| \right) \right\}\cdot\|\Delta^t_d\|  -  \|\Delta^t_d\|^2  ,
\end{aligned}
\end{equation}
}
where the third equality follow from $\mathcal{Q}_d^{t*} = W_d^{*} + \bar{\eta}^{t} 
 \nabla W_d^{*} = W_d^{*}$, where $\nabla W_d^{*} = 0$  by the optimal condition, and the first inequality follows from the contractivity of $\mathcal{Q}^t$ from Theorem~\ref{thm:contractivityG1wideN}.

Now we rethought \eqref{eq:mid}, by combining with \eqref{eq:mid2}, we obtain
\small{
\begin{equation}
\begin{aligned}
\label{eq:mid000}
&\mathbb{E}[\|\Delta^{t+1}_d\|^2]\\
\leq&  \mathbb{E}[\|\Delta^t_d\|^2] + (\bar{\eta}^{t})^2\mathbb{E}[\| \nabla {\hat{W}}_d^t\|^2]+ 2\bar{\eta}^{t} \mathbb{E}[\left< \nabla {W}_d^t - \nabla {W}_d^{*},\Delta^t_d\right>]  \\
\leq& \mathbb{E}[\|\Delta^t_d\|^2]    +    (\bar{\eta}^{t})^2\mathbb{E} [\|{\nabla \hat{W}}_d^t\|^2]  + 2\mathbb{E} \left[ \left\{  (1 - \bar{\eta}^{t}\xi)\|\Delta_d^t\| + \bar{\eta}^{t}\gamma\left(\sum_{i=1}^{d-1}\|\Delta_i^{t+1}\|  +   \sum_{i=d+1}^{L}\|\Delta_i^t\| \right)  \right\}    \cdot  \|\Delta^t_d\|  -  \|\Delta^t_d\|^2  \right] \\
\leq& \mathbb{E}[\|\Delta^t_d\|^2] + (\bar{\eta}^{t})^2\sigma_{d}^2 +  2\mathbb{E}\left[ \left\{ (1 - \bar{\eta}^{t}\xi)\|\Delta_d^t\|  +  \bar{\eta}^{t}\gamma\left(\sum_{i=1}^{d-1}\|\Delta_i^{t+1}\|  +     \sum_{i=d+1}^{L}\|\Delta_i^t\| \right)  \right\}  \cdot  \|\Delta^t_d\|  -  \|\Delta^t_d\|^2 \right],
\end{aligned}
\end{equation}
}
where $\sigma_{d}^2 = \sup_{\substack{W_1 \in B_2(r_1,W_1^{*}) \\ \dots \\ W_L \in B_2(r_L,W_L^{*})}}\mathbb{E}[\|\nabla_d \mathcal{J}^1(W_1,W_2,\dots,W_L)\|^2]$. 

After rearranging the above terms, we can transform \eqref{eq:mid000} to following inequality:
\begin{equation}
\begin{aligned}
\label{eq:mid3}
&\mathbb{E}[\|\Delta^{t+1}_d\|^2]   \\
\leq& (\bar{\eta}^{t})^2\sigma_{d}^2 + (1 - 2\bar{\eta}^{t}\xi)\mathbb{E}[\|\Delta^t_d\|^2]+ 2\bar{\eta}^{t}\gamma\mathbb{E} \left[ \left(\sum_{i=1}^{d-1}\|\Delta_i^{t+1}\|  +     \sum_{i=d+1}^{L}\|\Delta_i^t\| \right)  \|\Delta^t_d\| \right]  ,  \\
\leq& (\bar{\eta}^{t})^2\sigma_{d}^2 + (1-2\bar{\eta}^{t}\xi)\mathbb{E}[\|\Delta^t_d\|^2]  + \bar{\eta}^{t}\gamma\mathbb{E} \left[\sum_{i=1}^{d-1} \left(\|\Delta_i^{t+1}\|^2 + \|\Delta^t_d\|^2\right) \right] \\ 
& \quad + \bar{\eta}^{t}\gamma\mathbb{E} \left[\sum_{i=d+1}^{L} \left(\|\Delta_i^{t}\|^2 + \|\Delta^t_d\|^2\right)  \right]  \\
=& (\bar{\eta}^{t})^2\sigma_{d}^2 + \left[1 - 2\bar{\eta}^{t}\xi  +  \bar{\eta}^{t}\gamma(L-1) \right] \cdot \mathbb{E}[\|\Delta^t_d\|^2]  + \bar{\eta}^{t}\gamma\mathbb{E} \left[\sum_{i=1}^{d-1}\|\Delta_i^{t+1}\|^2\right]    \\  
& \quad +  \bar{\eta}^{t}\gamma\mathbb{E} \left[\sum_{i=d+1}^{L} \|\Delta_i^{t}\|^2\right]  ,  \\
\end{aligned}
\end{equation}
where the second inequality follows by by applying $2ab \leq a^2 + b^2$, so we have $ 2\| \Delta_i^{t} \| \| \Delta_d^t \| \leq \| \Delta_i^{t} \|^2 + \| \Delta_d^t \|^2 $, and the last equality follows by collecting like terms involving $\| \Delta_d^t \|^2$. According to \eqref{eq:mid3}, we can re-group the terms as follows

$$ \mathbb{E}[\|\Delta^{t+1}_d\|^2] - \bar{\eta}^{t}\gamma\mathbb{E} \left[\sum_{i=1}^{d-1}\|\Delta_i^{t+1}\|^2\right]  \leq  [1-2\bar{\eta}^{t}\xi + \bar{\eta}^{t}\gamma(L-1)] \cdot \mathbb{E}[\|\Delta^t_d\|^2]+ \bar{\eta}^{t}\gamma\mathbb{E} \left[\sum_{i=d+1}^{L} \|\Delta_i^{t}\|^2\right] + (\bar{\eta}^{t})^2\sigma_{d}^2  , $$
by summing the above inequality over $d$ from $1$ to $L$, we obtain

\begin{equation}
    \label{eq:mid4}
    \begin{aligned}
      &\mathbb{E}\left[\sum_{d=1}^L\|\Delta^{t+1}_d\|^2\right] - \bar{\eta}^{t}\gamma \mathbb{E} \left[\sum_{d=1}^L\sum_{i=1}^{d-1}\|\Delta_i^{t+1}\|^2\right]  \\
      \leq& [1 - 2\bar{\eta}^{t}\xi  +  \bar{\eta}^{t}\gamma(L - 1)]  \cdot \mathbb{E} \left[\sum_{d=1}^L\|\Delta^t_d\|^2\right]+ \bar{\eta}^{t}\gamma\mathbb{E} \left[\sum_{d=1}^L\sum_{i=d+1}^{L} \|\Delta_i^{t}\|^2\right]  +  (\bar{\eta}^{t})^2 \sum_{d=1}^L \sigma_{d}^2   \\  
      =& [1  +  \bar{\eta}^{t}\gamma(L - 1)]  \cdot \mathbb{E} \left[\sum_{d=1}^L\|\Delta^t_d\|^2\right] + [ - 2\bar{\eta}^{t}\xi] \cdot \mathbb{E} \left[\sum_{d=1}^L\|\Delta^t_d\|^2\right] + \bar{\eta}^{t}\gamma\mathbb{E} \left[\sum_{d=1}^L\sum_{i=d+1}^{L} \|\Delta_i^{t}\|^2\right]  +  (\bar{\eta}^{t})^2 \sum_{d=1}^L \sigma_{d}^2   , \\
      =&  [1 - 2\bar{\eta}^{t}\xi  +  \bar{\eta}^{t}\gamma(L - 1)]\mathbb{E}\left[\sum_{d=1}^L\|\Delta^t_d\|^2\right]+ \bar{\eta}^{t}\gamma\mathbb{E} \left[\sum_{d=1}^L\sum_{i=d+1}^{L} \|\Delta_i^{t}\|^2\right]  +  (\bar{\eta}^{t})^2\sigma^2 \\
      \leq& [1 - 2\bar{\eta}^{t}\xi  +  \bar{\eta}^{t}\gamma(L - 1)]\mathbb{E}\left[\sum_{d=1}^L\|\Delta^t_d\|^2\right]+ \bar{\eta}^{t}\gamma(L - 1)\mathbb{E} \left[\sum_{d=1}^L \|\Delta_d^{t}\|^2\right]+ (\bar{\eta}^{t})^2\sigma^2  .
    \end{aligned}
\end{equation}
where $\sigma = \sqrt{\sum_{d=1}^L\sigma_{d}^2}$. Furthermore, it can be readily verified that the following equation holds:
\begin{align*}
\mathbb{E}\left[\sum_{d=1}^L\|\Delta^{t+1}_d\|^2\right] - \bar{\eta}^{t}\gamma(L-1)\mathbb{E} \left[\sum_{d=1}^L\|\Delta_d^{t+1}\|^2\right]\leq \mathbb{E}\left[\sum_{d=1}^L\|\Delta^{t+1}_d\|^2\right] - \bar{\eta}^{t}\gamma\mathbb{E} \left[\sum_{d=1}^L\sum_{i=1}^{d-1}\|\Delta_i^{t+1}\|^2\right]  .
\end{align*}
Therefore, rearranging the above formula, we have
\begin{align}
\label{eq:mid5}
\left[ 1- \bar{\eta}^{t}\gamma(L-1) \right] \mathbb{E} \left[\sum_{d=1}^L\|\Delta_d^{t+1}\|^2\right] \leq \mathbb{E}\left[\sum_{d=1}^L\|\Delta^{t+1}_d\|^2\right] - \bar{\eta}^{t}\gamma\mathbb{E} \left[\sum_{d=1}^L\sum_{i=1}^{d-1}\|\Delta_i^{t+1}\|^2\right]  .
\end{align}

In conclusion, combining \eqref{eq:mid4} and \eqref{eq:mid5}, we obtain:
\begin{align*}
& [1- \bar{\eta}^{t} \gamma (L-1) ]\mathbb{E}\left[\sum_{d=1}^L\|\Delta^{t+1}_d\|^2\right]\\
\leq &\mathbb{E}\left[\sum_{d=1}^L\|\Delta^{t+1}_d\|^2\right] - \bar{\eta}^{t}\gamma\mathbb{E} \left[\sum_{d=1}^L\sum_{i=1}^{d-1}\|\Delta_i^{t+1}\|^2\right] \\
\leq&   [1 - 2\bar{\eta}^{t}\xi  +  \bar{\eta}^{t}\gamma(L - 1)]\mathbb{E}\left[\sum_{d=1}^L\|\Delta^t_d\|^2\right]+ \bar{\eta}^{t}\gamma(L - 1)\mathbb{E} \left[\sum_{d=1}^L \|\Delta_d^{t}\|^2\right] + (\bar{\eta}^{t})^2\sigma^2 \\
=& [1 - 2\bar{\eta}^{t}\xi  +  2\bar{\eta}^{t}\gamma(L - 1)]\mathbb{E}\left[\sum_{d=1}^L\|\Delta^t_d\|^2\right]  +  (\bar{\eta}^{t})^2\sigma^2      ,
\end{align*}
which can rearrange the above formula:
\begin{align}
\label{eq:mid6}
\mathbb{E}\left[\sum_{d=1}^L\|\Delta^{t+1}_d\|^2\right] &\leq \frac{1 - 2\bar{\eta}^{t}\xi  +  2\bar{\eta}^{t}\gamma(L - 1)}{1-\bar{\eta}^{t}\gamma(L-1)}\mathbb{E}\left[\sum_{d=1}^L\|\Delta^t_d\|^2\right] +  \frac{(\bar{\eta}^{t})^2}{1 - \bar{\eta}^{t}\gamma(L - 1)}\sigma^2   ,
\end{align}
where $\gamma < \frac{2\xi}{3(L-1)}$, and $\frac{1 - 2\bar{\eta}^{t}\xi  +  2\bar{\eta}^{t}\gamma(L - 1)}{1-(L-1)\bar{\eta}^{t}\gamma} < 1$.

\subsubsection*{Part II: Proof of Theorem~\ref{theorem:final}}
\label{proof:part2}
To obtain the final theorem~\ref{theorem:final}, we need to expand the recursion from Part \hyperref[proof:part1]{I} of the proof of Theorem~\ref{theorem:final}. We rearrange \eqref{eq:mid6}, then obtain

\begin{align*}
&\mathbb{E}\left[\sum_{d=1}^L\|\Delta^{t+1}_d\|^2\right] \\
\leq& \frac{1 - 2\bar{\eta}^{t}[\xi  -  \gamma(L - 1)]}{1-\bar{\eta}^{t}\gamma(L-1)}\mathbb{E}\left[\sum_{d=1}^L\|\Delta^t_d\|^2\right]+ \frac{(\bar{\eta}^{t})^2}{1 - \bar{\eta}^{t}\gamma(L - 1)}\sigma^2   \\
=& \left(1  -  \frac{\bar{\eta}^{t}[2\xi - 3\gamma(L - 1)]}{1 - \bar{\eta}^{t}\gamma(L - 1)}\right)\mathbb{E}\left[\sum_{d=1}^L\|\Delta^t_d\|^2\right] + \frac{(\bar{\eta}^{t})^2}{1 - \bar{\eta}^{t}\gamma(L - 1)}\sigma^2      .
\end{align*}
To simplify the process of proof, we defined $p^t$ and $s^{t}$ as
\begin{equation}
\label{eq:define_ps}
\begin{aligned}
    &p^t = 1 - \frac{1-2\bar{\eta}^{t}\xi+2\bar{\eta}^{t}\gamma(L-1)}{1-\bar{\eta}^{t}\gamma(L-1)} = \frac{\bar{\eta}^{t}[2\xi - 3\gamma(L - 1)]}{1 - \bar{\eta}^{t}\gamma(L - 1)}   \\
    &s^t = \frac{(\bar{\eta}^{t})^2}{1 - \bar{\eta}^{t}\gamma(L-1)}  .
\end{aligned}
\end{equation}

Thus we have
\begin{equation}
\begin{aligned}
\label{eq:t2-mid1}
&\mathbb{E}\left[\sum_{d=1}^L\|\Delta^{t+1}_d\|^2\right]  \\
\leq&      (1-p^t)\mathbb{E}\left[\sum_{d=1}^L\|\Delta^t_d\|^2\right]  +  s^t\sigma^2\\
\leq&    (1-p^t)\left\{(1-p^{t-1})\mathbb{E}\left[\sum_{d=1}^L\|\Delta^{t-1}_d\|^2\right]  +  s^{t-1}\sigma^2\right\}+     s^t\sigma^2\\
=&    (1 - p^t)(1 - p^{t-1})\mathbb{E}\left[\sum_{d=1}^L\|\Delta^{t-1}_d\|^2\right]  +  (1-p^t)s^{t-1}\sigma^2  +s^t\sigma^2\\
\leq&      (1 - p^t)(1 - p^{t-1}) \left\{ (1 - p^{t-2})\mathbb{E} \left[\sum_{d=1}^L \|\Delta^{t-2}_d\|^2\right]   +  s^{t-2}\sigma^2 \right\} + (1-p^t)s^{t-1}\sigma^2  +  s^t\sigma^2\\
=&  (1 - p^t)(1 - p^{t-1})(1 - p^{t-2})\mathbb{E}\left[\sum_{d=1}^L\|\Delta^{t-2}_d\|^2\right] + (1 - p^t)(1 - p^{t-1})s^{t-2}\sigma^2   +  (1 - p^t)s^{t-1} \sigma^2  +  s^t\sigma^2    . \\
\end{aligned}
\end{equation}
By applying mathematical induction, following from \eqref{eq:t2-mid1}, we obtain the following:
\begin{equation}
    \label{eq:final}
    \mathbb{E}\left[\sum_{d=1}^L\|\Delta^{t+1}_d\|^2\right]  \leq 
    \underbrace{ \mathbb{E}\left[\sum_{d=1}^L\|\Delta^0_d\|^2\right]\prod_{i=0}^t(1 - p^i) }_{\text{I}}+ \underbrace{  \sigma^2\sum_{i=0}^{t-1} s^i\prod_{j=i+1}^t(1 - p^j) }_{\text{II}} + \underbrace{ s^t\sigma^2 }_{\text{III}}. 
\end{equation}
According to the Assumption \ref{assu:bounder} of $\bar{\eta}^{t}$, when $B_u = \frac{1}{ \gamma (L-1)} $, so we have 
$$0 < B_l \leq \bar{\eta}_{\text{min}} \leq \bar{\eta}^{t} \leq \bar{\eta}_{\text{max}} < B_u = \frac{1}{\gamma(L-1) } < 1,$$
therefore, by the definition \eqref{eq:define_ps}, we can derive that the denominator $1 - \bar{\eta}^{t} \gamma(L-1)  > 0$ concerning $p^{t}$ and $s^{t}$. Assuming that $\gamma < \frac{2\xi}{3(L-1)}$, so there exists $2\xi-3\gamma(L-1)>0$, let $\delta := \gamma(L - 1)$, we have
\begin{align*}
p^t = \frac{\bar{\eta}^{t} (2\xi - 3\delta)}{1 - \delta \bar{\eta}^{t}} \Rightarrow p_{\text{min}} := \frac{\bar{\eta}_{\text{min}} (2\xi - 3\delta)}{1 - \delta \bar{\eta}_{\text{min}}}, \quad \text{and} \quad p_{\text{max}} := \frac{\bar{\eta}_{\text{max}} (2\xi - 3\delta)}{1 - \delta \bar{\eta}_{\text{max}}}   ,
\end{align*}
and 
\begin{align*}
s^t = \frac{(\bar{\eta}^{t})^2}{1 - \delta \bar{\eta}^{t}} \Rightarrow s_{\max} := \frac{(\bar{\eta}_{\max})^2}{1 - \delta \bar{\eta}_{\max}}   ,
\end{align*}
where the above two formula follows from the monotonicity of $p^{t}$ and $s^{t}$ .

Now we start to analyze for \eqref{eq:final}, according to the property of $p_{\min}$, $p_{\max}$, and $s_{\max}$, firstly, for the first term \text{I} on the right side of the whole inequality, we have
\begin{align*}
\prod_{i=0}^t (1-p^i) \leq (1-p_{\min})^{t+1} = \exp \left( - (t+1) \log \left( \frac{1}{1-p_{\min}} \right) \right) = e^{-c(t+1)},
\end{align*}
where $ c := \log \left( \frac{1}{1-p_{\min}} \right) > 0$. Hence, we have 
\begin{align}
    \label{eq:final1}
    \mathbb{E} \left[ \sum_{d=1}^L \| \Delta_d^0 \|^2 \right] \prod_{i=0}^t (1 - p^i) \leq C_0 e^{-c(t+1)}   ,
\end{align}
where $C_{0} = \mathbb{E} \left[ \sum_{d=1}^L \| \Delta_d^0 \|^2 \right] $ denotes a constant. Secondly, for the second term \text{II} on the right side of \eqref{eq:final}, we have
\begin{align*}
\sum_{i=0}^{t-1} s^i \prod_{j=i-1}^{t} (1 - p^j)  \leq s_{\max} \sum_{i=0}^{t-1} \prod_{j=i-1}^{t} (1 - p^j) ,
\end{align*}
due to
$$\prod_{j=i+1}^t (1-p^j) \leq (1-p_{\min})^{t-i} = e^{-c(t-i)} \Rightarrow \sum_{i=0}^{t-1} e^{-c(t-i)} = e^{-ct} \sum_{i=0}^{t-1} e^{ci} = e^{-ct} \cdot \frac{e^{ct} - 1}{e^c - 1} , $$ therefore, we have
\begin{equation}
    \label{eq:final2}
    \sum_{i=0}^{t-1} s^i \prod_{j=i+1}^{t} (1 - p^j) \leq s_{\max} \cdot \frac{1 - e^{-c t}}{1 - e^{-c}} \leq s_{\max} \cdot \frac{1}{1 - e^{-c}}  ,
\end{equation}
let $B = s_{\max} \cdot \frac{1}{1 - e^{-c}}$. Finally, because $B$ is a constant value, for the third term \text{III} for \eqref{eq:final}, according to $\bar{\eta}^{t} \leq \bar{\eta}_{max} \leq \frac{1}{\delta}$, we derive that
\begin{align}
  \label{eq:final3}
  s^t \leq s_{\max} \Rightarrow   \sigma^2 s^t \leq \sigma^2 s_{\max}   .
\end{align}
In conclusion, combining with \eqref{eq:final1}, \eqref{eq:final2}, and \eqref{eq:final3}, we have 

\begin{align*}
\mathbb{E} \left[ \sum_{d=1}^L \left\| \Delta_d^{t+1} \right\|^2 \right] \leq C_0 e^{-ct} + \sigma^2 B + \sigma^2 s_{\max}    ,
\end{align*}
or
\begin{align*}
\mathbb{E}\left[\sum_{d=1}^{L} \| \Delta_d^{t+1} \|^2 \right] \leq O(e^{-ct}) + O(1)   .
\end{align*}
This completes the proof of Theorem \ref{theorem:final}.


\subsection*{B. Additional Experiment Results} 

\subsection*{B.1 Sensitivity Analysis}
\label{sec:B.1}

\begin{figure}[H]
\centering
\includegraphics[width=0.88\linewidth]{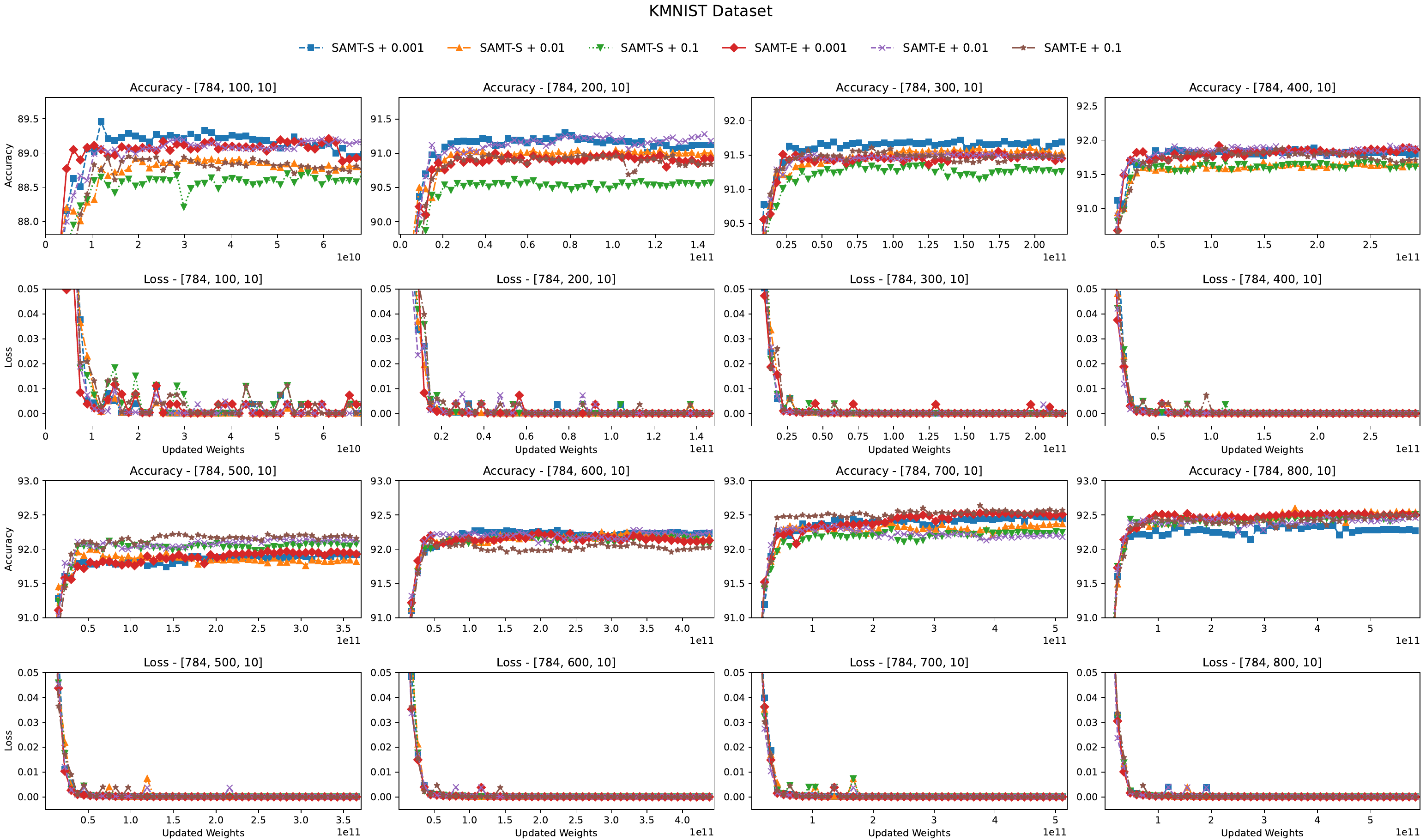}
    \caption{Different $\bar{\eta}^{0}$: Test accuracy and loss curves for KMNIST dataset. } 

\label{fig:mlp-kmnist-diffeta}
\end{figure}

On the KMNIST dataset, as shown in Figure \ref{fig:mlp-kmnist-diffeta}, the SAMT-E and SAMT-S methods demonstrate comparable advantages. In particular, SAMT-E tends to achieve superior performance when using higher initial step sizes, while SAMT-S exhibits better performance under lower initial step sizes.

\subsection*{B.2 Ablation Studies}
\label{sec:B.2}

\begin{figure}[H]
\centering
\includegraphics[width=0.88 \linewidth]{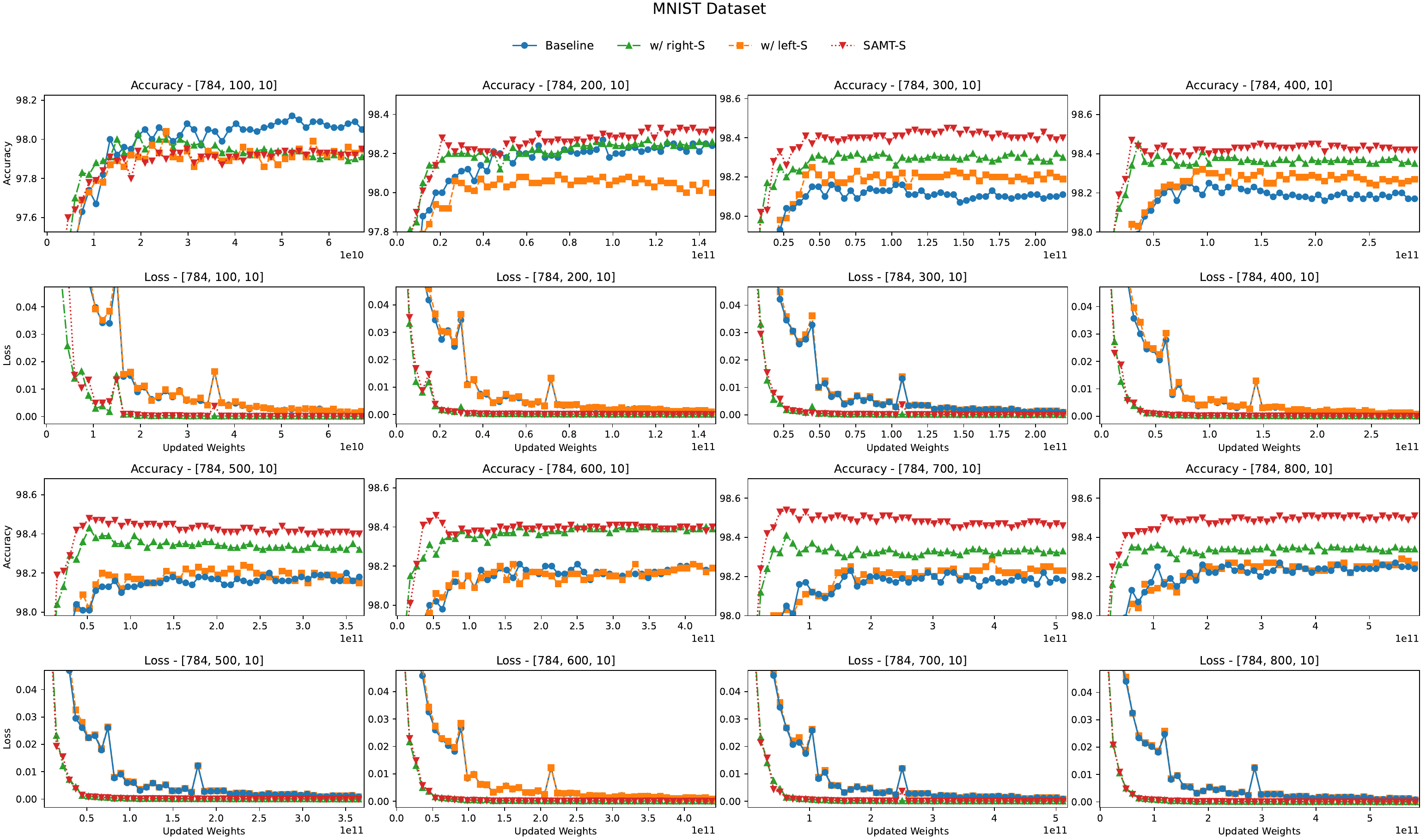}
    \caption{Ablation experiments for SAMT-S method: Test accuracy and loss curves for MNIST dataset. } 

\label{fig:mlp-mnist-abstudy-s}
\end{figure}

\begin{figure}[H]
\centering
\includegraphics[width=0.88\linewidth]{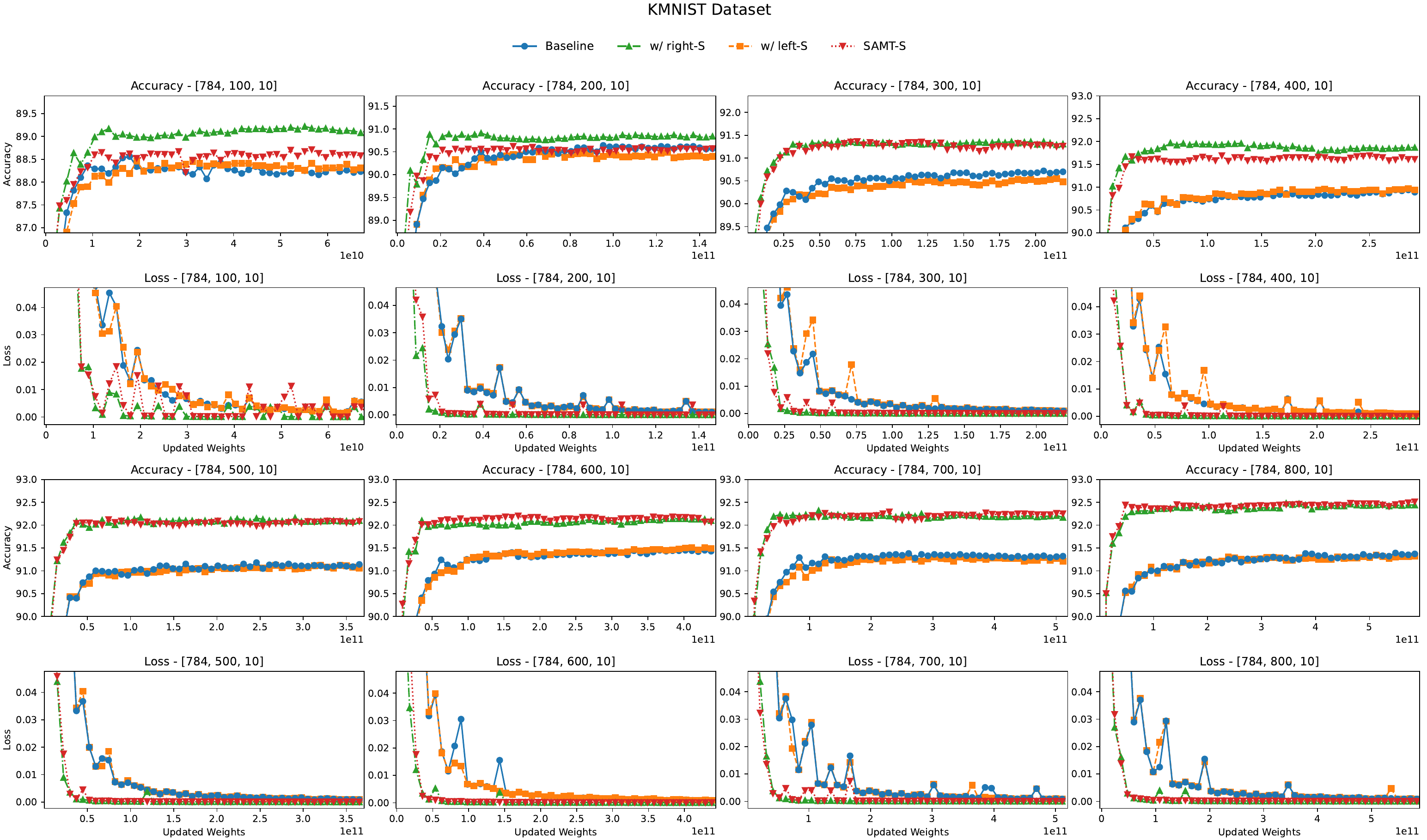}
    \caption{Ablation experiments for SAMT-S method: Test accuracy and loss curves for KMNIST dataset. } 

\label{fig:mlp-kmnist-abstudy-s}
\end{figure}

Figures \ref{fig:mlp-mnist-abstudy-s} and \ref{fig:mlp-kmnist-abstudy-s} present the ablation experiments for the SAMT-S method on different datasets. From the trends observed in the curves, it is evident that, when compared to the baseline algorithm with a fixed scalar step size and the $\beta \cdot \bar{\eta}^{0}$ variant in \eqref{eq:update_etann} that utilizes only the left term, the temporary step sizes $(1 - \beta) \cdot \hat{\eta}$ (only utilizes the right term) generated by our proposed eta model $\psi$ outperform these two ablation baselines in most dimensions. Furthermore, the complete SAMT-S method exhibits enhanced competitiveness in high-dimensional neural network architectures.




\begin{figure}[H]
\centering
\includegraphics[width=0.88 \linewidth]{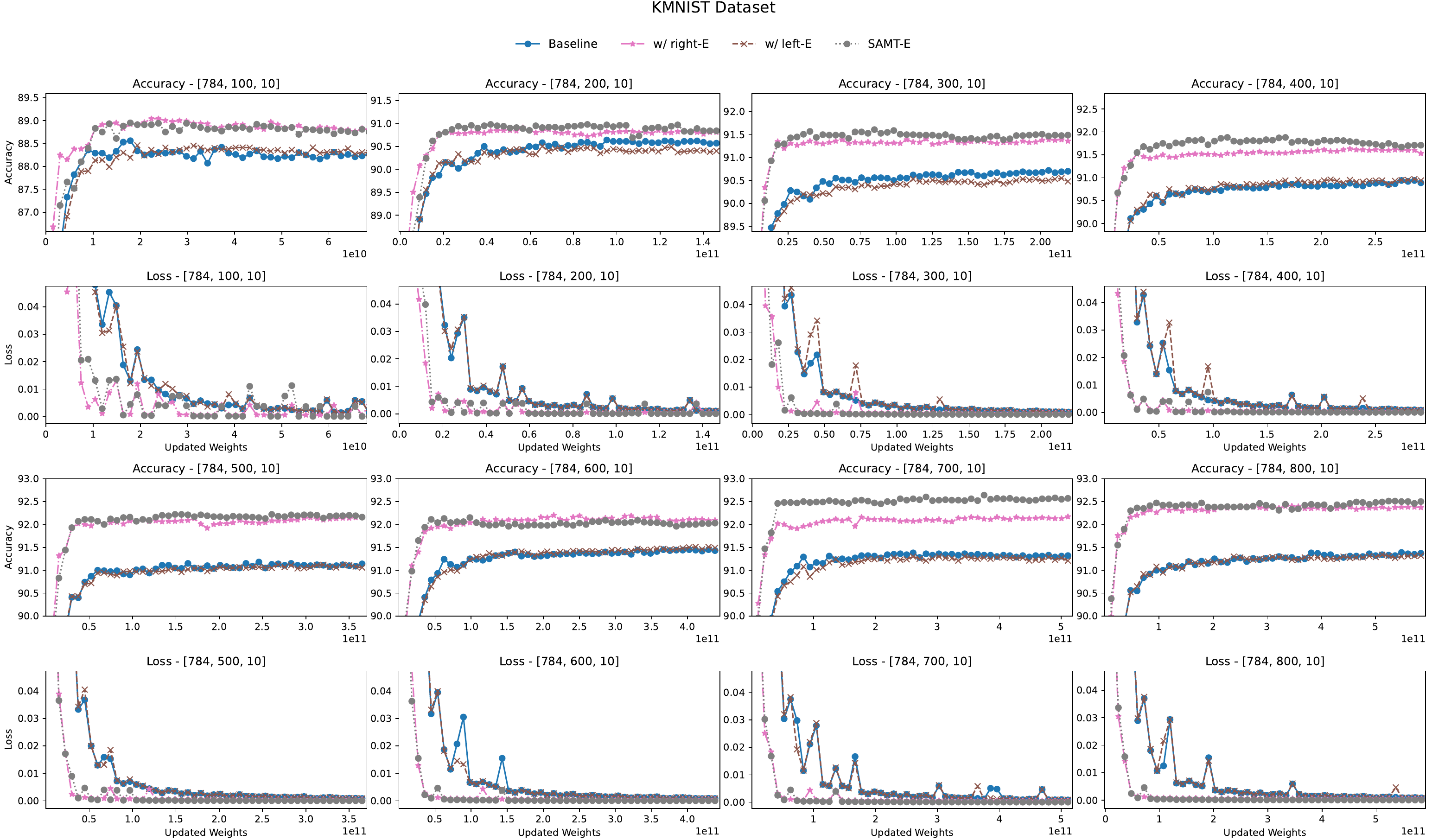}
    \caption{Ablation experiments for SAMT-E method: Test accuracy and loss curves for KMNIST dataset. } 

\label{fig:mlp-kmnist-abstudy-e}
\end{figure}

In the ablation experiment on the KMNIST dataset, as shown in Figure \ref{fig:mlp-kmnist-abstudy-e}, the advantage of the temporary step sizes $\beta \odot \hat{\eta}$ generated by the eta model $\psi$ can also be observed. Furthermore, after incorporating the initial step size, the complete SAMT-E method demonstrates more balanced performance across different settings.

\subsection*{B.3 Different Projection Styles: Tanh vs Sigmoid} 
\label{sec:B.3}
\begin{figure}[H] 
\centering
\includegraphics[width=0.88\linewidth]{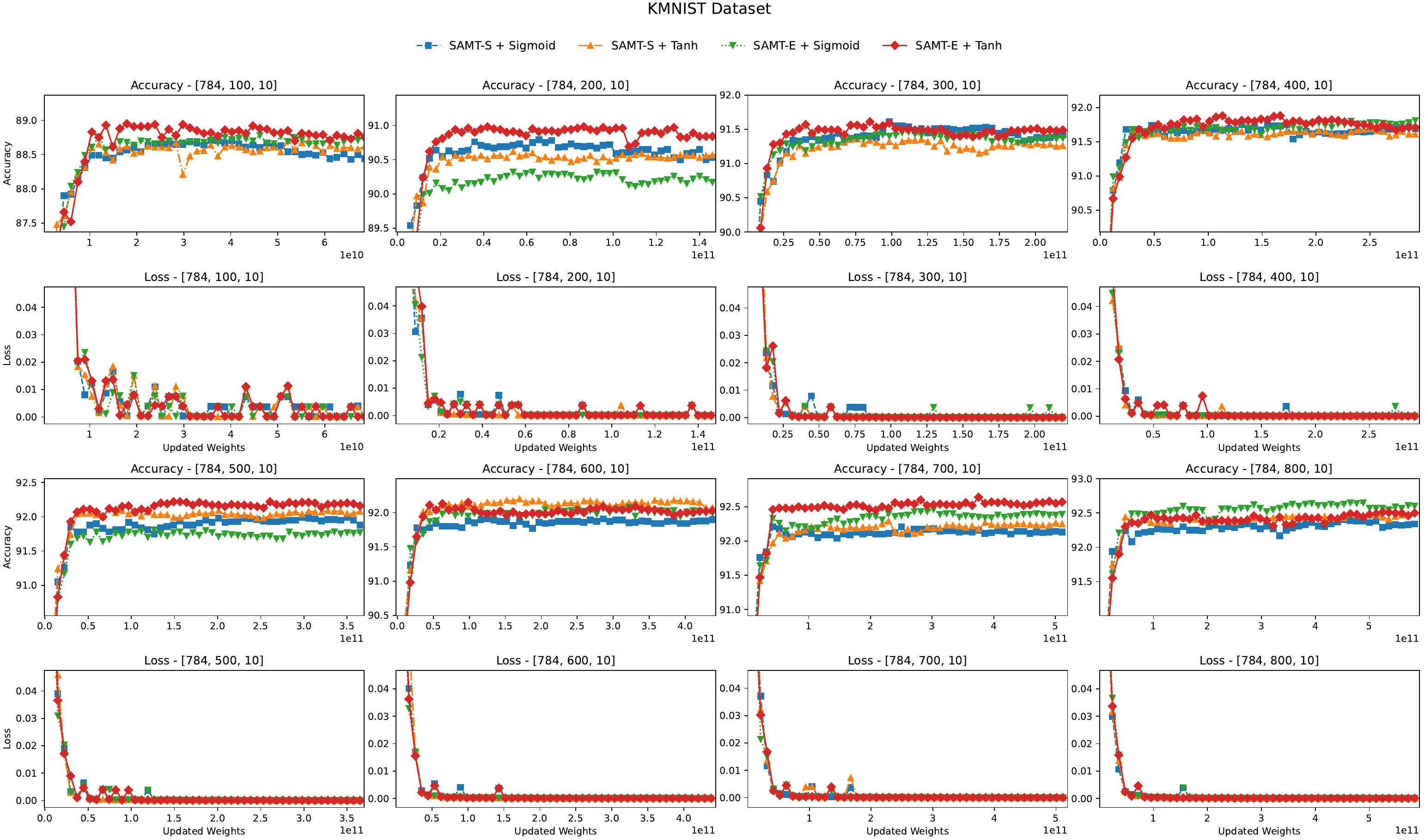}
    \caption{Different projection style: Tanh vs Sigmoid: Test accuracy and loss curves for KMNIST dataset. } 

\label{fig:mlp-kmnistvs}
\end{figure}

On the KMNIST dataset, as shown in Figure \ref{fig:mlp-kmnistvs}, the SAMT-E method with the Tanh projection consistently outperforms other combined algorithms across the vast majority of neural network dimensions, further demonstrating the advantages conferred by the high dynamic range of the Tanh function. For the SAMT-S method, the Sigmoid projection slightly outperforms the Tanh projection on low-dimensional neural networks; however, on high-dimensional structures, the Tanh-based method exhibits superior performance.









\bibliographystyle{plain}   


\end{document}